\documentclass[letterpaper]{article} 
\usepackage{aaai25}  
\usepackage{times}  
\usepackage{helvet}  
\usepackage{courier}  
\usepackage[hyphens]{url}  
\usepackage{natbib}  
\usepackage{caption} 
\usepackage{url}
\usepackage{graphicx} 

\usepackage{microtype}
\usepackage{subfigure}
\usepackage{algpseudocode}

\usepackage{amssymb}
\usepackage{epstopdf, color, array}
\usepackage{epsfig}  
\usepackage{transparent}
\usepackage{amsmath,bm}
\usepackage{mathrsfs} 
\usepackage[english]{babel}
\usepackage{blindtext}
\usepackage{amsfonts}%
\usepackage{amsthm}%
\usepackage{mathtools}
\usepackage{mathrsfs}%
\usepackage[title]{appendix}%
\usepackage{xcolor}%
\usepackage{textcomp}%
\usepackage{manyfoot}%
\usepackage{adjustbox}
\usepackage{listings}%

\newcommand{\mathbbm}[1]{\text{\usefont{U}{bbm}{m}{n}#1}}

\usepackage{algorithm}

\usepackage{newfloat}
\usepackage{listings}
\DeclareCaptionStyle{ruled}{labelfont=normalfont,labelsep=colon,strut=off} 
\floatstyle{ruled}
\newfloat{listing}{tb}{lst}{}
\floatname{listing}{Listing}

\title{Learning Hidden Subgoals under Temporal Ordering Constraints in Reinforcement Learning}

\author{
    Duo Xu,
    Faramarz Fekri    
}
\affiliations{
    School of ECE \\
    Georgia Institute of Technology
    \\

    225 North Avenue, Atlanta, GA 30332 USA\\
    dxu301@gatech.edu
}

\usepackage{bibentry}

\begin{document}

\maketitle

\begin{abstract}
In real-world applications, the success of completing a task is often determined by multiple key steps which are distant in time steps and have to be achieved in a fixed time order. For example, the key steps listed on the cooking recipe should be achieved one-by-one in the right time order. These key steps can be regarded as subgoals of the task and their time orderings are described as temporal ordering constraints. 
However, in many real-world problems, subgoals or key states are often hidden in the state space and their temporal ordering constraints are also unknown, which make it challenging for previous RL algorithms to solve this kind of tasks.
In order to address this issue, in this work we propose a novel RL algorithm for {\bf l}earning hidden {\bf s}ubgoals under {\bf t}emporal {\bf o}rdering {\bf c}onstraints (LSTOC). 
We propose a new contrastive learning objective which can effectively learn hidden subgoals (key states) and their temporal orderings at the same time, based on first-occupancy representation and temporal geometric sampling. In addition, we propose a sample-efficient learning strategy to discover subgoals one-by-one following their temporal order constraints by building a subgoal tree to represent discovered subgoals and their temporal ordering relationships. Specifically, this tree can be used to improve the sample efficiency of trajectory collection, fasten the task solving and generalize to unseen tasks.
The LSTOC framework is evaluated on several environments with image-based observations, showing its significant improvement over baseline methods. \footnote{Source codes will be released upon acceptance}
\end{abstract}

\section{Introduction}
\label{sec:introduction}
In real life, successfully completing a task often involves multiple temporally extended key steps, where these key steps have to be achieved in specified time orders. For instance, in the process of making chemicals, different operations have to be strictly performed in the right time order, e.g., sulfuric acid must be added after water. Otherwise, the right chemical reaction can never occur or even the safety will be threatened. These key steps are necessary for the success of completing the given task and skipping any of them or doing them in the wrong time order will lead to failure of the task. In this work, these key steps are regarded as subgoals.
Tasks consisting of multiple subgoals with temporal ordering constraints are common in many real-world applications, such as the temporal logic tasks in control systems and robotics \citep{baier2008principles}. Since these tasks may have long-time horizon and sparse reward, the knowledge of subgoals and their temporal orderings are necessary for modern RL algorithms to solve these tasks efficiently. 
However, these subgoals can be hidden and unknown in many real-world scenarios. For instance, due to the user's lack of knowledge, these subgoals may be missing when specifying the task. Alternatively, due to the partial observability of environment, the agent does not know subgoals and their temporal orderings in advance. 

\begin{figure}    
    \vspace{10pt}
    \centering
    \subfigure[Layout]{
        \centering
        \includegraphics[width=1.in]{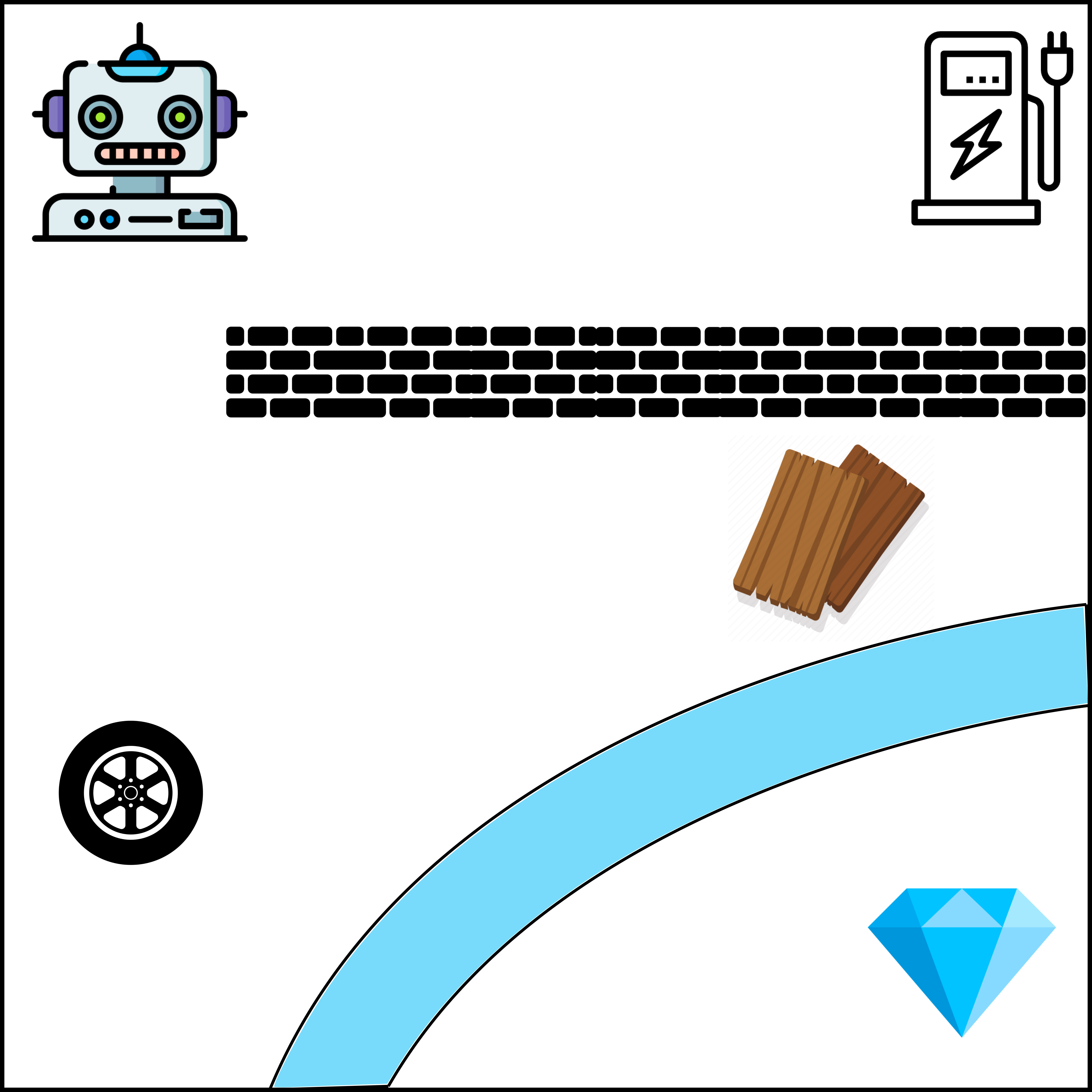}
        \label{fig:tl_example_0}
    }
    \subfigure[FSM]{
        \centering
        \fontsize{6pt}{10pt}\selectfont
        \def\svgwidth{1.1in}
        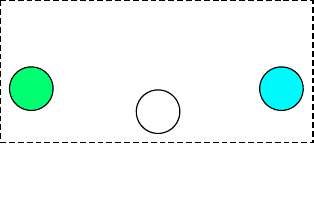
        \label{fig:tl_example_1}
    }
    \vspace{-10pt}
    \caption{ (a) Example task. (b) The FSM for temporal dependencies of subgoals. Letters "c", "b", "w" and "d" are short for charger, board, wheel and diamond, respectively.}
    \label{fig:tl_example}
    \vspace{-20pt}
\end{figure}

\noindent
{\bf Motivating example.} For example, consider a service robot tasked to collect the diamond in limited time steps, as shown in Figure \ref{fig:tl_example_0}. Due to the limited power and the blockage of river, the agent has to first go to the charger to get charged, then pick up wheel or board to go across the river, and finally get the diamond. If the agent first picks up the wheel or board and then goes to the charger, the task cannot be finished in the required time steps. The temporal dependencies of these subgoals can be described by the finite state machine (FSM) in Figure \ref{fig:tl_example_1}. The temporal logic language for describing these dependencies is c;(b$\vee$w);d. However, since the agent can only observe things around him, it is not aware of the river and does not know that charger, wheel and board are subgoals, i.e., subgoals are hidden to the agent.  

When solving tasks with hidden subgoals, the binary label at the end of the episode, indicating the task is accomplished successfully or not, is the only reward information for the agent to leverage to solve the task. This existing situation can be challenging for modern RL algorithms which use Bellman equation to propagate value estimates back to the earlier key steps \citep{sutton2018reinforcement}. These algorithms suffer from slow convergence and expensive learning complexity, which are going to be verified by empirical experiments. Therefore, it is necessary to develop new RL algorithms to solve the task which contains multiple {\it hidden} subgoals with {\it unknown} temporal ordering constraints. To the best knowledge, this is the first work which investigates this problem.

In this work, we propose a novel framework for {\bf L}earning hidden {\bf S}ubgoals under {\bf T}emporal {\bf O}rdering {\bf C}onstraints in RL (LSTOC). It consists of the learning subgoal and the labeling components. In learning subgoals, the proposed framework efficiently discovers states or observations corresponding to hidden subgoals and learns their temporal dependencies by using contrastive learning, which iteratively builds a subgoal tree (denoted as $\mathcal{T}$ and defined in Section \ref{sec:tl}) by discovered subgoals to represent learned temporal ordering constraints. Specifically, in $\mathcal{T}$, nodes are labeled with discovered key states of subgoals and edges represent their temporal ordering relationships. This tree $\mathcal{T}$ is used to guide the trajectory collection, ground the semantic meaning of learned subgoals (labeling component), accelerate the task solving and help the generalization. 



\noindent
{\bf Subgoal Learning.} In order to improve the sample efficiency, we propose a new learning method which discovers hidden subgoals one-by-one and builds $\mathcal{T}$ iteratively representing discovered subgoals and their temporal ordering relationships. The trajectory collection is guided by $\mathcal{T}$ to focus more on the working node $v_w$ which is a node in $\mathcal{T}$ not fully explored yet. In every iteration of building $\mathcal{T}$, by using a novel contrastive learning method, the agent only discovers the {\it next} subgoal which is next to the subgoal of working node $v_w$ under temporal ordering constraints, and expands $\mathcal{T}$ by adding this newly discovered subgoal as a new child to $v_w$. Then, this new child node will be used as working node, initiating the next iteration of tree expansion. 
This iterating process will stop whenever the success of every collected trajectory about task completion can be explained by the constructed $\mathcal{T}$, meaning that $\mathcal{T}$ is fully constructed, i.e., all the hidden subgoals and temporal orderings have been learned in $\mathcal{T}$. 

\noindent
{\bf Contrastive Learning.} In order to discovery subgoals next to working node $v_w$, we propose a new contrastive learning method. In this case, since only the {\it first} visit to the next subgoal is meaningful, based on pre-processed trajectories, the proposed method first computes the {\it first-occupancy representation} (FR) \citep{moskovitz2021first} of trajectories by removing repetitive states. Then, we will use contrastive learning to detect subgoals. However, since conventional contrastive learning could detect multiple subgoals without giving any temporal distance information and the next subgoal is temporally closest to $v_w$ among detected ones, the real next subgoal we need to discover could be missed by using conventional methods. Therefore, it is necessary to learn the temporal distances (i.e., temporal orderings) of detected subgoals and then select the temporally closest one as next subgoal. Therefore, we propose a new contrastive learning objective which can detect key states for subgoals and learn their temporal orderings at the same time. It formulates the contrastive learning objective by using {\it temporal geometric sampling} to sample one positive state from the FR of a processed positive trajectory and several negative states from the FR of a batch of processed negative trajectories. 
To be best of our knowledge, we are the first to propose a contrastive learning method which can detects key states and learn their temporal distances at the same time.

\noindent
{\bf Labeling.} In the labeling part, if the specification representing the temporal dependencies of subgoal semantic symbols is given, we formulate an integer linear programming (ILP) problem to determine the mapping from the discovered key states of subgoals to subgoal semantic symbols, making every path of the constructed subgoal tree $\mathcal{T}$ satisfy the given specification. When this ILP problem is solved, the labeling function is obtained, essentially giving semantic meaning to every learned subgoal.

We evaluate LSTOC on $9$ tasks in three environments, including Letter, Office, and Crafter domains. In these environments, with image-based observations, the agent needs to visit different objects under temporal ordering constraints. Our evaluations show that LSTOC can outperform baselines on learning subgoals and efficiency of solving given tasks. The generalizability of LSTOC is also empirically verified. The limitation of LSTOC is discussed finally.

\section{Related Works}
\label{sec:related}
Recently linear temporal logic (LTL) formulas have been widely used in Reinforcement Learning (RL) to specify temporal logic tasks \citep{littman2017environment}. Some papers develop RL algorithms to solve tasks in the LTL specification \citep{camacho2019ltl,de2019foundations,bozkurt2020control}.
In some other papers, authors focus on learning the task machine from traces of symbolic observations based on binary labels received from the environment \citep{gaon2020reinforcement,xu2021active,ronca2022markov}.
However, all these papers assume the access to a labeling function which maps raw states into propositional symbols, working in the labeled MDP \citep{hasanbeig2019reinforcement}. 

There are some papers assuming to have an imperfect labeling function, where the predicted symbols can be erroneous or uncertain \citep{li2022noisy,hatanaka2023reinforcement}. 
But these papers do not address the problem of learning subgoals. A recent paper studies the problem of grounding LTL in finite traces (LTLf) formulas in image sequences \citep{umili2023grounding}. However, their method is only applicable to offline cases with static dataset and does not consider the online exploration in the environment. In addition, authors in \citep{luo2023learning} propose an algorithm for learning rational subgoals based on dynamic programming. However, Their approach requires the availability of the state transition model, which is not feasible in general real-world applications.

Contrastive learning was used to detect subgoals (key states) in previous RL papers \cite{zhang2023learning,sun2023contrastive,casper2023open,park2022surf,liang2022reward}. However, these methods sample clips of positive and negative trajectories to formulate the contrastive objective. It can make the temporal distances of states not distinguishable and cannot learn temporal distances of detected subgoals. Therefore, these methods cannot be used to learn subgoals under temporal ordering constraints. 

\section{Preliminaries}
\label{sec:prelim}

\subsection{Reinforcement Learning}
\label{sec:rl}
Reinforcement learning (RL) is a framework for learning the strategy of selecting actions in an environment in order to maximize the collected rewards over time \citep{sutton2018reinforcement}. The problems addressed by RL can be formalized as Markov decision processes (MDP), defined as a tuple $\mathcal{M}=\langle\mathcal{S}, \mathcal{A}, \mathcal{T}, R, \gamma, \mathcal{S}_0\rangle$, where $\mathcal{S}$ is a finite set of environment states, $\mathcal{A}$ is a finite set of agent actions, $\mathcal{T}:\mathcal{S}\times\mathcal{A}\times\mathcal{S}\to[0,1]$ is a probabilistic transition function, $R:\mathcal{S}\times\mathcal{A}\to[R_{\text{min}}, R_{\text{max}}]$ is a reward function with $R_{\text{min}}, R_{\text{max}}\in\mathbb{R}$ and $\gamma\in[0,1)$ is a discount factor. Note that $\mathcal{S}_0$ is the set of initial states where the agent starts in every episode, and $S_0:s_0\sim\mathcal{S}_0$ is a distribution of initial states. 

In addition, corresponding to the MDP, we assume that there is a set of semantic symbols $\mathcal{G}$ representing subgoals. We also define a {\it labeling function} $L:\mathcal{S}\to\mathcal{G}\cup\{\varnothing\}$ that maps an environmental state to a subgoal which is a key step to accomplish the task. Furthermore, the {\it key state} is defined as the environmental state whose output of function $L$ is a subgoal in $\mathcal{G}$. Most states of the environment are not key states where the outputs of $L$ are empty ($\varnothing$). In this work, the states corresponding to subgoals and the labeling function are all unknown to the agent initially. The agent can only leverage collected trajectories and their labels of task accomplishing results to solve the task.

\subsection{Temporal Logic Language}
\label{sec:tl}
We assume that the temporal dependencies (orderings) of subgoals considered in this work can be described by a formal language $\mathcal{TL}$ composed by three operators. Syntactically, all subgoals in $\mathcal{G}$ are in $\mathcal{TL}$, and $\forall \varphi_1, \varphi_2\in\mathcal{TL}$, the expressions $(\varphi_1;\varphi_2)$, $(\varphi_1\vee \varphi_2)$ and $(\varphi_1\wedge \varphi_2)$ are all in $\mathcal{TL}$, representing "$\varphi_1$ then $\varphi_2$", "$\varphi_1$ or $\varphi_2$" and "$\varphi_1$ and $\varphi_2$", respectively. 
Formally, a trajectory of states $\tau=(s_1,\ldots,s_n)$ satisfies a task description $\varphi$, written as $\tau\models\varphi$, whenever one of the following holds:
\begin{itemize}
    \item If $\varphi$ is a single subgoal $g\in\mathcal{G}$, then the first state of $\tau$ must not satisfy $g$, and instead the last state must satisfy $g$, which implies that $\tau$ has at least 2 states
    \item If $\varphi=(\varphi_1;\varphi_2)$, then $\exists0<j<n$ such that $(s_1,\ldots,s_j)\models\varphi_1$ and $(s_j,\ldots,s_n)\models\varphi_2$, i.e., task $\varphi_1$ should be finished before $\varphi_2$
    \item If $\varphi=(\varphi_1\vee\varphi_2)$, then $\tau\models\varphi_1$ or $\tau\models\varphi_2$, i.e., the agent should either finish $\varphi_1$ or $\varphi_2$
    \item If $\varphi=(\varphi_1\wedge\varphi_2)$, then $\tau\models(\varphi_1;\varphi_2)$ or $\tau\models(\varphi_2;\varphi_1)$, i.e., the agent should finish both $\varphi_1$ and $\varphi_2$ in any order
\end{itemize}
Note that the language $\mathcal{TL}$ for specifying temporal dependencies is expressive enough and  covers LTLf \citep{de2013linear} which is a finite fragment of LTL without using "always" operator $\Box$.
\begin{figure}
    \vspace{10pt}
    \centering
    \fontsize{6pt}{10pt}\selectfont
    \def\svgwidth{2.5in}
    \def\svgheight{1.2in}
    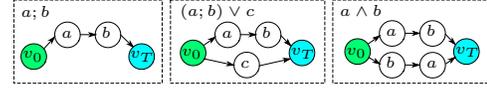
    \caption{ Examples of TL formulas and their corresponding FSMs. The initial node is $v_0$ and the accepting (terminal) node is $v_T$.}
    \label{fig:fsm}
    \vspace{-10pt}
\end{figure}

Every task specification $\varphi\in\mathcal{TL}$ can be represented by a non-deterministic finite-state machine (FSM) \citep{luo2023learning}, representing the temporal orderings and branching structures. Each FSM $\mathcal{M}_{\varphi}$ of task $\varphi$ is a tuple $(V_{\varphi}, E_{\varphi}, I_{\varphi}, F_{\varphi})$ which denote subgoal nodes, edges, the set of initial nodes and the set of accepting (terminal) nodes, respectively. 
Some examples of FSMs are shown in Figure \ref{fig:fsm}. 
There exists a deterministic algorithm for transforming any specification in $\mathcal{TL}$ to a unique FSM \citep{luo2023learning}. 


\begin{figure}
    \vspace{5pt}
    \centering
    \fontsize{6pt}{7pt}\selectfont
    \def\svgwidth{1.6in}
    \def\svgheight{1.in}
    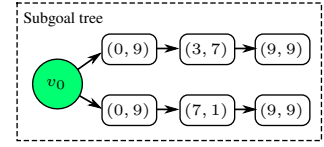
    \caption{ The subgoal tree representing the subgoal temporal dependencies $c;(b\vee w);d$. 
    }
    \label{fig:fsm_tree}
    \vspace{-10pt}
\end{figure}

First, we define the {\it satisfying sequence} as the sequence of key states which can satisfy the task. Second, the tree formed by all the satisfying sequences of the task is defined as the {\it subgoal tree}. For example, for the problem in Figure \ref{fig:tl_example_0}, its satisfying sequences are $[(0,9), (3,7), (9,9)]$ and $[(0,9), (7,1), (9,9)]$ ($(x,y)$ denotes the coordinates of state), where $(0,9), (3,7), (7,1)$ and $(9,9)$ represent states corresponding to subgoals "c", "b", "w" and "d", respectively. Its subgoal tree is shown in Figure \ref{fig:fsm_tree}. 
In this work, we only consider the subgoal temporal dependencies whose FSMs do not have any loops. Hence, an FSM in this category can be converted into a tree. 
Note that subgoals and their temporal dependencies (i.e., ordering constraints) are hidden and unknown to the agent. The algorithm proposed in this work is to learn these subgoals and their temporal ordering constraints based on collected trajectories and labels whether the task is accomplished or not.

\subsection{First-occupancy Representation}
\label{sec:prelim_fr}
We use first-occupancy representation (FR) for learning subgoals. FR measures the duration in which a policy is expected to reach a state for {\it the first time}, which emphasizes the first occupancy.\\
\noindent
{\bf Definition 1.}\citep{moskovitz2021first} For an MDP with finite $\mathcal{S}$, the first-occupancy representation (FR) for a policy $\pi$ $F^{\pi}\in[0,1]^{|\mathcal{S}|\times|\mathcal{S}|}$ is given by
\begin{equation}
    F^{\pi}(s,s'):=\mathbb{E}_{\pi}\bigg[\sum_{k=0}^{\infty}\gamma^k\mathbbm{1}(s_{t+k}=s',s'\not\in\{s_{t:t+k}\})\bigg|s_t=s\bigg] \label{fr}
\end{equation}
where $\{s_{t:t+k}\}=\{s_t,s_{t+1},\ldots,s_{t+k-1}\}$ and $\{s_{t:t+0}\}=\emptyset$. The above indicator function $\mathbbm{1}$ equals 1 only when $s'$ first occurs at time $t+k$ since time $t$. So $F^{\pi}(s,s')$ gives the expected discount at the time the policy first reaches $s'$ starting from $s$. 

\subsection{Contrastive Learning in RL}
\label{sec:prelim_contrastive_learning}
Contrastive learning was used in previous RL algorithms to detect important states \cite{sun2023contrastive,zhang2023learning}, which learns an importance function $f_{\omega}$ by comparing positive and negative trajectories. The objective loss function can be written as
\begin{small}
\begin{eqnarray}
\lefteqn{\mathcal{L}(\omega;\mathcal{B}_P, \mathcal{B}_N)=} \nonumber \\
&&
-\mathbb{E}_{\substack{\tau_p\sim\mathcal{B}_P,\\ \tau_n\sim\mathcal{B}_N}}\bigg[\frac{\sum_{\nu=1}^L\exp(f_{\omega}(\tau_p[\nu]))}{\sum_{\nu=1}^L\exp(f_{\omega}(\tau_p[\nu])) + \sum_{\mu=1}^L\exp(f_{\omega}(\tau_n[\mu]))}\bigg] \label{old_contrastive_obj}
\end{eqnarray}
\end{small}
which aims making important states to have high value at the output of $f_{\omega}$. However, this conventional method has two major shortcomings. First, it cannot learn temporal distances of important states. In our work, hidden subgoals are under temporal ordering constraints, and different subgoals may have different temporal distances away from the initial state. But function $f_{\omega}$ trained here treated every subgoal equally, which cannot detect the real next subgoal to achieve. Second, the numerator in \eqref{old_contrastive_obj} contains multiple states and the importance value ($f_{\omega}$ output) of real subgoal may not stand out among its neighboring states, which can make the subgoal learning inaccurate. In order to tackle these two issues, we propose a new contrastive learning method.

\section{Methodology}
\label{sec:method}
In this work, we propose the LSTOC framework for learning hidden subgoals and their temporal ordering constraints by leveraging trajectories and results of task accomplishment (positive or negative labels) collected from the environment. Based on constructed subgoal tree, the agent can solve the task faster and generalize to other unseen tasks involving same subgoals.

\begin{figure*}[ht]
    \centering
    \fontsize{9pt}{10pt}\selectfont
    \def\svgwidth{4.2in}
    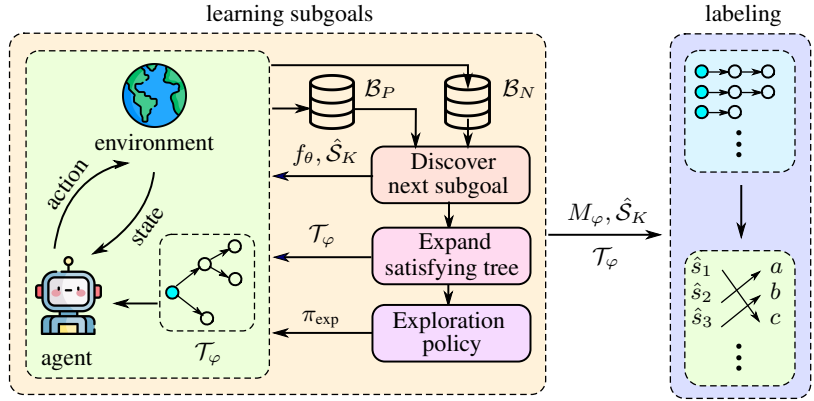
    \caption{ Diagram of the LSTOC framework. For learning subgoals, $\mathcal{B}_P$ ($\mathcal{B}_N$) represents the buffer of positive (negative) trajectories, $f_{\theta}$ is the state representation function, $\hat{\mathcal{S}}_K$ is the set of discovered key states, ${\mathcal{T}}_{\varphi}$ is the subgoal tree, and $\pi_{\text{exp}}$ is the exploration policy. The trajectory collection is guided by ${\mathcal{T}}_{\varphi}$ and $\pi_{\text{exp}}$. In the labeling part, based on $\mathcal{M}_{\varphi}$, $\hat{\mathcal{S}}_K$ and ${\mathcal{T}}_{\varphi}$, the mapping from discovered key states to subgoal symbols is determined by solving an ILP problem, yielding the labeling function. $\mathcal{M}_{\varphi}$ denotes the FSM of temporal dependencies of subgoals in task $\varphi$.
    }
    \label{fig:ground_diagram}
\end{figure*}

\subsection{General Context}
\label{sec:general} 

The diagram of LSTOC is shown in Figure \ref{fig:ground_diagram}. In learning subgoal, the agent iteratively learns hidden subgoals one-by-one in a depth first manner and builds a subgoal tree $\mathcal{T}_{\varphi}$ to guide the trajectory collection and represent learned subgoals and temporal orderings. For example, as the problem shown in Figure \ref{fig:tl_example}, the agent will first learn subgoal "c", then "w", then "d", and then "b", finally "d", building the subgoal tree in Figure \ref{fig:fsm_tree}.

In every iteration, the agent focuses on learning subgoals next to the current working node by using a contrastive learning method, and expands $\mathcal{T}_{\varphi}$ by adding a new leaf node labeled with the newly learned subgoal, which is used as the working node in the next iteration. This iteration will be repeated until the success of every trajectory on task completion can be explained by the learned subgoals and their temporal relationships on $\mathcal{T}_{\varphi}$. Then, the agent will proceed to the labeling component.
When collecting a trajectory $\tau_k$, by using an exploration policy $\pi_{\text{exp}}$ trained with reward shaping method, the agent is guided by $\mathcal{T}_{\varphi}$ to reach the working node (an unexplored node of $\mathcal{T}_{\varphi}$). A binary label $l_k$ indicating the result of task completion is received at the end of $\tau_k$. Positive ($l_k=1$) and negative ($l_k=0$) trajectories are stored into positive ($\mathcal{B}_P$) and negative buffers ($\mathcal{B}_N$), respectively. 

Once the FSM $\mathcal{M}_{\varphi}$ which describes subgoal temporal dependencies by semantic symbols in $\mathcal{G}$ becomes available, the labeling component of LSTOC can determine the mapping from discovered key states to subgoal symbols in $\mathcal{G}$ by solving an integer linear programming (ILP) problem, leveraging $\mathcal{T}_{\varphi}$ to impose the semantic meaning onto every learned subgoal. 


\noindent
{\bf Subgoal Notation.} We define the set $\hat{\mathcal{S}}_K$ as an ordered set of discovered key states which form a subset of the state space of the environment, i.e., $\hat{\mathcal{S}}_K\subset\mathcal{S}$. Every node of the subgoal tree is labeled by a state in $\hat{\mathcal{S}}_K$. 
For the $k$-th state in $\hat{\mathcal{S}}_K$, i.e., $\hat{s}_k$, $k$ is the index for indicating {\it detected subgoal} and $\hat{s}_k$ is the {\it key state} corresponding to the detected subgoal $k$. Only newly discovered key state not included in $\hat{\mathcal{S}}_K$ will be added to $\hat{\mathcal{S}}_K$, creating an index indicating a newly detected subgoal.

\noindent
{\bf Remark.} Although subgoals are hidden, the result of the task completion can still be returned by the environment to the agent for every trajectory. This is common in robotic applications. For example, in service robot, when the task specification misses key steps, the robot can learn to discover these missing steps based on its trajectory data and feedbacks about task completion from users or other external sources \citep{gonzalez2021service}.


\subsection{Learning Subgoals}
\label{sec:learning_subgoal}
Since the task is assumed to have multiple temporally extended hidden subgoals, it is impractical to collect sufficient informative trajectories for an offline supervised learning approach to detect all the subgoals at once. Therefore, we propose a new learning method to discover subgoals one-by-one iteratively.
In every iteration, in addition to trajectory collection, the agent conduct operations including discovering next subgoal, expanding the subgoal tree and training the exploration policy, which will be introduced as following. 

\begin{figure*}
    \centering
    \fontsize{6pt}{10pt}\selectfont
    \def\svgwidth{5.3in}
    \def\svgheight{1.2in}
    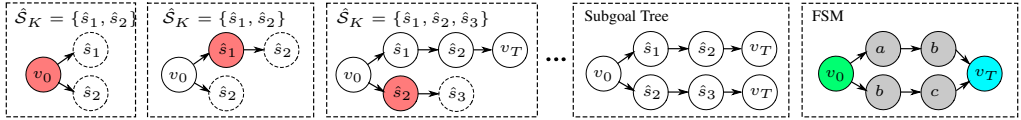
    \caption{ Examples of building subgoal tree ${\mathcal{T}}_{\varphi}$. The temporal dependencies of subgoals can be expressed as $(a;b)\vee(b;c)$, whose FSM is shown in the rightmost figure. In the left three figures, the red node denotes the working node $v_w$, and $\hat{\mathcal{S}}_K$ is given on the upper left corner. The dashed nodes are unexplored nodes to be explored. Every node is labeled with a discovered key state and its index. The fourth figure shows a fully built ${\mathcal{T}}_{\varphi}$. The subgoals of the task are hidden and the agent only knows the result of task completion for each episode. 
    }
    \label{fig:fsm_exp}
    \vspace{-10pt}
\end{figure*}

\subsubsection{Expand Subgoal Tree}
\label{sec:reconstruct}

{\bf Tree Definition.} As discussed in Section \ref{sec:tl}, the temporal ordering constraints of discovered subgoals can be expressed by the subgoal tree ${\mathcal{T}}_{\varphi}$. In ${\mathcal{T}}_{\varphi}$, except the root, each node $v_n$ is labeled by a key state $\hat{s}_{k_{v_n}}$ of the $k_{v_n}$-th learned subgoal in $\hat{\mathcal{S}}_K$. Specifically, in ${\mathcal{T}}_{\varphi}$, the children of a node labeled with subgoal $p$ contain key states of next subgoals to achieve after $p$ is visited in the FSM of subgoal temporal dependencies. Since children of every node are only labeled by the temporally nearest subgoals for task completion, the temporal orderings of subgoals can be well represented in ${\mathcal{T}}_{\varphi}$. Whenever $\mathcal{T}_{\varphi}$ is well established, every path from the root to a leaf node in ${\mathcal{T}}_{\varphi}$ corresponds to a satisfying sequence of the task (concept introduced in Section \ref{sec:tl}).

\begin{figure}
    \centering
    \fontsize{6pt}{10pt}\selectfont
    \def\svgwidth{1.6in}
    \def\svgheight{1.2in}
    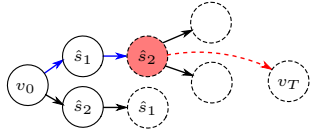
    \caption{ A trajectory conditioned on $v_w$. The red node is the working node $v_w$. The dashed nodes are unexplored nodes. Blue: visits the path $\xi_w:=[\hat{s}_1, \hat{s}_2]$. Red: explores till the end of the episode. Only the red part is used to discover key states of next subgoals.}
    \label{fig:traj_exp}
    \vspace{-10pt}
\end{figure}

\noindent
{\bf Tree Expansion.} Based on discovered key states, ${\mathcal{T}}_{\varphi}$ is expanded iteratively in a depth-first manner. 
Initially, ${\mathcal{T}}_{\varphi}$ only has the root node $v_0$. For a node $v_l$, we define the {\it path} of $v_l$ (denoted as $\xi_l$) as the sequence of key states along the path from $v_0$ to $v_l$ in ${\mathcal{T}}_{\varphi}$, as an example shown in Figure \ref{fig:traj_exp}. We first define nodes whose paths has not lead to task completion yet as {\it unexplored} nodes. In each iteration of tree expansion, the agent first selects an unexplored node $v_w$ from ${\mathcal{T}}_{\varphi}$ as the {\it working node}, and then focuses on discovering next subgoals to achieve after visiting $v_w$ in $\mathcal{T}_{\varphi}$. Examples of building the subgoal tree are shown in Figure \ref{fig:fsm_exp}, where the dashed nodes are unexplored nodes.

\noindent
{\bf Expansion at a Working Node.} For expansion at $v_w$, the agent first explores the collection of trajectories conditioned on $v_w$ and then discovers next subgoals by using contrastive learning. We define a trajectory which visits key states along the {\it path} of $v_w$ (i.e., $\xi_w$) sequentially as a trajectory {\it conditioned on} the working node $v_w$. 
An example of trajectory conditioned on $v_w$ is shown in Figure \ref{fig:traj_exp}, where the path $\xi_w$ is $[\hat{s}_1, \hat{s}_2]$.
Every trajectory is collected by applying the exploration policy $\pi_{\text{exp}}$. In order to encourage the agent to collect trajectories conditioned on $v_w$, $\pi_{\text{exp}}$ is trained by reward shaping method guided by $\mathcal{T}_{\varphi}$, which is introduced in Section \ref{sec:exp_policy}.

If the number of collected positive trajectories conditioned on $v_w$ is larger than a threshold $N_T$, then the agent initiates its process of subgoal discovery at working node $v_w$. Specifically, the agent picks up trajectories conditioned on $v_w$ from $\mathcal{B}_P$ and $\mathcal{B}_N$, and only keeps the part after $v_w$ is visited in every trajectory (red part in Figure \ref{fig:traj_exp}). Based on these pre-processed trajectories, the key states of next subgoals to visit (after $v_w$) are discovered by using contrastive learning method (introduced in Section \ref{sec:discover}). Then, the newly discovered key state is used to build a new node $v_{\text{new}}$, which is added to ${\mathcal{T}}_{\varphi}$ as a new child of $v_w$. The new working node will be at $v_{\text{new}}$, initiating a new iteration of tree expansion at $v_{\text{new}}$. Since the agent is not familiar with the environment initially, we design an adaptive mechanism of selecting $N_T$, which is introduced in Section \ref{sec:algorithm}.


\noindent
{\bf Satisfying Sequence.} At working node $v_w$, if the agent can complete the task successfully whenever he follows the path $\xi_w$, then the path $\xi_w$ is regarded as a satisfying sequence of the task (concept introduced in Section \ref{sec:tl}). 
In this case, we say that positive trajectories which complete the task successfully by following the path $\xi_w$ are {\it explained} by the path $\xi_w$. The path $\xi_w$ will be added into the set of discovered satisfying sequences $\mathcal{P}_S$. Then, the node $v_w$ becomes fully explored and the parent node $v_p$ of $v_w$ will be selected as the new working node, initiating a new iteration of expansion at $v_p$.

\noindent
{\bf Discovering Alternative Branches.}
With working node at $v_p$, as long as we can find positive trajectories conditioned on $v_p$ and not explained by any discovered satisfying sequences in $\mathcal{P}_S$, there must be alternative hidden subgoals next to $v_p$ which are not discovered yet. The agent will continue exploring and discovering hidden subgoals (next to $v_p$) based on these unexplained trajectories, until the success of every positive trajectory conditioned on $v_p$ is explained by descendent nodes of $v_p$ in $\mathcal{T}_{\varphi}$ (i.e., $v_p$ becomes fully explored). Specifically, if unexplained trajectories conditioned on $v_p$ are not sufficient, the agent applies $\pi_{\text{exp}}$ to explore for collecting more trajectories, until the number of unexplained positive trajectories conditioned at $v_p$ is larger than $N_T$. If new key states of next subgoals are found, these states will be used to build new nodes added to $\mathcal{T}_{\varphi}$ as new children of $v_p$, and additional iterations of expansion at these new nodes will be initiated. Otherwise, $v_p$ becomes fully explored, and the parent of $v_p$ will be selected as the new working node.

\noindent
{\bf Termination Condition of Tree Expansion.} The tree expansion will stop only when 1) every collected positive trajectory can be explained by $\mathcal{T}_{\varphi}$ or 2) $\mathcal{T}_{\varphi}$ becomes structurally inconsistent with $M_{\varphi}$ (the FSM for subgoal temporal dependencies). 
Condition 1) means that every positive trajectory $\tau_p$ can be explained by a satisfying sequence $\xi_l$ which is a path in $\mathcal{T}_{\varphi}$. Condition 2) means that the current $\mathcal{T}_{\varphi}$ is wrong and has to be built from the scratch again. The details of condition 2) are discussed in Appendix \ref{sec:app_algo}. If condition 2) is true, the LSTOC framework will be reset and restart again with a larger threshold $N_T$.

\subsubsection{Discovering Next Subgoal}
\label{sec:discover}
In each iteration of LSTOC framework, based on the up-to-date subgoal tree $\mathcal{T}_{\varphi}$, the agent focuses on learning next subgoal to achieve after visiting {\it working node} (concept introduced in Section \ref{sec:reconstruct}). For instance, in the example of Figure \ref{fig:tl_example}, if the up-to-date $\mathcal{T}_{\varphi}$ has two nodes labeled by key states of learned subgoals "c" and "b" ("c" is the parent of "b") and the working node is at "b", then the next subgoal to discover is the state of "d". 
As discussed in Section \ref{sec:prelim_contrastive_learning}, although contrastive learning was used in RL to detect important states, there are some issues making it inapplicable to solve our problem.
In this work, we propose a new contrastive learning method to discover key state of next subgoal by introducing some new techniques.


\noindent
{\bf First-occupancy Representation (FR).} Starting from the working node $v_w$ of $\mathcal{T}_{\varphi}$, only the first visit to the next subgoal is meaningful for completing the task. In tasks with multiple subgoals under temporal ordering constraints, repetitive state visitations, especially those to non-key states, are frequent and can distract the weight function $f_{\omega}$ from paying attention to real key states. Therefore, we propose to compute and use the FRs of trajectories to conduct contrastive learning. Specifically, for any trajectory $\tau$ {\it conditioned on} $v_w$ (an example shown in Figure \ref{fig:traj_exp}), we select the part of $\tau$ after $v_w$ is visited in $\mathcal{T}_{\varphi}$ and remove repetitive states there, where only the first visit to any state is kept, yielding the FR of $\tau$, denoted as $\tilde{\tau}$.

\noindent
{\bf Remark. } Comparison of states is needed when removing repetitive states in a trajectory. However, the raw state or observation may not be directly comparable. In this case, we propose to learn a distinguishable $d$-dimensional representation of any state or observation in buffer $\mathcal{B}$, i.e., $o_{\theta}(\cdot):\mathcal{S}\to\mathbb{R}^d$, based on the contrastive loss in \cite{oord2018representation}. Then, if two states $s_1$ and $s_2$ have high cosine similarity of vectors $o_{\theta}(s_1)$ and $o_{\theta}(s_2)$, i.e., greater than $0.99$, states $s_1$ and $s_2$ will be regarded as the same.

\noindent
{\bf Contrastive Learning Objective.} Since the task has multiple temporally extended hidden subgoals, applying conventional contrastive learning to train $f_{\omega}$ based on FRs of selected trajectories would produce multiple subgoals, some of which are not next subgoal to achieve. For example shown in Figure \ref{fig:tl_example}, when $\mathcal{T}_{\varphi}$ only has one node labeled with the key state of subgoal "c" and working node $v_w$ is also at that node, direct application of conventional contrastive learning can produce multiple key states including states of "b", "w" and "d". However, "d" is not the next subgoal to achieve after achieving "c", which can distract the learning of hidden subgoals. {\it Since hidden subgoals need to be achieved following temporal ordering constraints, the agent needs to select only one of the next subgoals to expand the subgoal tree $\mathcal{T}_{\varphi}$, which has the smallest temporal distance away from subgoal of the working node $v_w$.} 

Therefore, we propose a new contrastive learning objective which can detect key states for subgoals and learn their temporal distances at the same time. The proposed contrastive objective is formulated by one positive sample and multiple negative ones. Specifically, for any trajectory $\tau^+$ conditioned on $v_w$ randomly selected from $\mathcal{B}_P$, with the FR of pre-processed trajectory  (introduced above) denoted as $\tilde{\tau}^+$, the positive sample is to sample one state $s_{t^+}$ from $\tilde{\tau}^+$ where the time index $t^+$ is sampled from the temporal geometric distribution, i.e., $t^+\sim\text{Geom}(1-\gamma)$\footnote{$\text{Geom}(1-\gamma)$ is the geometric distribution with parameter $1-\gamma$.} and $\gamma$ is the discount factor of the environmental MDP. The negative samples have $B$ states, denoted as $\{s_{t_i^-}\}_{i=1}^B$. To sample each $s_{t_i^-}$, a negative trajectory $\tau_i^-$ conditioned on $v_w$ is randomly selected from $\mathcal{B}_N$. After pre-processing and computing the FR of $\tau_i^-$, $s_{t_i^-}$ is sampled from $\tilde{\tau}_i^-$ with $t_i^-\sim\text{Geom}(1-\gamma)$. Therefore, the proposed contrastive learning objective can be formulated as below, 

\begin{small}
\begin{eqnarray}
\lefteqn{\mathcal{L}_{\text{contrastive}}(\omega; \{s_{t^+}\}, \{s_{t_i^-}\}_{i=1}^B)=} \nonumber \\
&&\frac{\exp(f_{\omega}(o_{\theta}(s_{t^+})))}{\exp(f_{\omega}(o_{\theta}(s_{t^+}))))+\sum_{i=1}^B\exp(f_{\omega}(o_{\theta}(s_{t^-_i})))} \label{con-obj}
\end{eqnarray}
\end{small}

In the objective function above, the numerator is a function of only one state $s_{t^+}$ sampled from a positive trajectory while the denominator has multiple states sampled from negative trajectories. This can make the importance function $f_{\omega}$ assign high value to the right state more concentratedly, without being distracted by neighbors of the real key state. The temporal geometric sampling can make states, which are temporally closer to the initial states of any trajectories, have higher chances of being used to formulate the contrastive learning objective. This can make the value of $f_{\omega}$ to be inversely proportional to the temporal distance of the input state, so that the temporal ordering can be reflected by the values of $f_{\omega}$, i.e., the higher the value of $f_{\omega}$, the higher the temporal order will be .

After the objective in \eqref{con-obj} is maximized with sufficient number of states sampled from positive and negative trajectories, the outputs of $f_{\omega}$ at key states of subgoals can be much larger than other states. The state with highest value of $f_{\omega}$ (highest temporal order) will be selected as the key state of next subgoal to achieve and be used to expand the subgoal tree $\mathcal{T}_{\varphi}$.

\subsubsection{Exploration Policy}
\label{sec:exp_policy}
The exploration policy $\pi_{\text{exp}}$ is realized by a GRU-based policy model. Each action is history dependent and drawn from the action distribution at the output of the policy model. 
The policy $\pi_{\text{exp}}$ is trained by the recurrent PPO algorithm \citep{goyal2020recurrent,ni2021recurrent} which extends the classical PPO \citep{schulman2017proximal} into POMDP domain. The binary label of task satisfaction given at the end of the trajectory is used as a reward for training $\pi_{\text{exp}}$. 

Additionally, in order to guide the exploration to collect more trajectories conditioned on the working node $v_w$ in $\mathcal{T}_{\varphi}$, auxiliary rewards ($r_a>0$) will be given to visits to the key states along the path $\xi_w$ from root to $v_w$ in the right temporal order specified by path $\xi_w$. Training $\pi_{\text{exp}}$ with these rewards can encourage the agent to collect more trajectory data which are useful for discovering next subgoals at $v_w$. For the example in Figure \ref{fig:traj_exp}, if any collected trajectory $\tau$ is conditioned on $v_w(=\hat{s}_2)$, $r_a$ will be given to the first visit to $\hat{s}_1$ and then the first visit to $\hat{s}_2$. If $\hat{s}_2$ is first visited in a trajectory $\tau'$, no $r_a$ will be given to states in $\tau'$, since $\tau'$ follows another path on $\mathcal{T}_{\varphi}$ not containing $v_w$ and hence is not conditioned on $v_w$.

\subsection{Labeling}
\label{sec:labeling}
Given the constructed subgoal tree ${\mathcal{T}}_{\varphi}$ and FSM $\mathcal{M}_{\varphi}$, in the labeling part of LSTOC, the mapping from $\hat{\mathcal{S}}_K$ to subgoal semantic symbols $\mathcal{G}$ (labeling function $L_{\varphi}$) is determined by solving an integer linear programming (ILP) problem, which makes every discovered satisfying sequence lead to an accepting state in $\mathcal{M}_{\varphi}$ and hence yields the semantic meaning of every hidden subgoal. The details of the ILP problem are introduced in Appendix \ref{sec:app_labeling}

\subsection{Algorithm}
\label{sec:algorithm}
Since the agent knows little about the environment initially, it is difficult to set the initiation condition (the value $N_T$) of subgoal discovery at every working node. If $N_T$ is too small, there would not be enough trajectory data for subgoal discovery. If $N_T$ is too large, too many trajectories will be redundant, reducing the learning efficiency. Therefore, we propose an adaptive mechanism to try different values of $N_T$ from small to large. Specifically, there is a main loop (Algorithm \ref{alg:main} in Appendix \ref{sec:app_algo}), trying to call LSTOC framework with $N_T=K_T,2K_T,\ldots$, until hidden subgoals are correctly learned and the ILP problem is successfully solved (Line 8 and 9 of Algorithm \ref{alg:lstoc} in Appendix \ref{sec:app_algo}), yielding the labeling function $L_{\varphi}$. The tables of algorithms are presented in Appendix \ref{sec:app_algo}.

\section{Experiments}
\label{sec:experiments}
The experiments aim at answering the following questions: 
\begin{enumerate}
    \item How well the proposed contrastive learning method would detect key states for subgoals and learn their temporal distances? 
    \item Can we solve the task and learn hidden subgoals more efficiently with the help of building the subgoal tree?
    \item Would the learned subgoals help the generalization to other new tasks?
\end{enumerate} 
The details of the environments are introduced in Appendix \ref{sec:env}. The tasks used for evaluation are presented in Appendix \ref{sec:tasks}. Then, experimental results answering questions 1) and 2) above will be presented in Section \ref{sec:con_perf} and \ref{sec:learning_efficiency}. Question 3) will be answered in Appendix \ref{sec:generalization}. Limitation of LSTOC will be discussed finally. Implementation details and more results are also included in Appendix.


\subsection{Results}
\label{sec:results}

\subsubsection{Learning Subgoals}
\label{sec:con_perf}
In this section, we focus on the evaluation of the proposed contrastive learning method which consists of two operations: 1) computing the first-occupancy representations (FR) of trajectories; 2) formulating the contrastive objective by using temporal geometric sampling to sample one state from every selected trajectory. The detailed introduction is in Section \ref{sec:discover}. 

According to these two operations, we design two baseline contrastive learning methods for comparison. In Baseline 1, the FR computation is omitted, and the formulation of contrastive learning objective is same as the one in LSTOC. In Baseline 2, the FR computation is kept the same, but the contrastive learning objective is formulated by the conventional method in \cite{sun2023contrastive,zhang2023learning}, where the objective is computed by every states in selected positive and negative trajectories, without specific state sampling process. 

\begin{figure}
    \centering
    \subfigure[Letter Task 1]{
        \centering
        \includegraphics[width=1.5in]{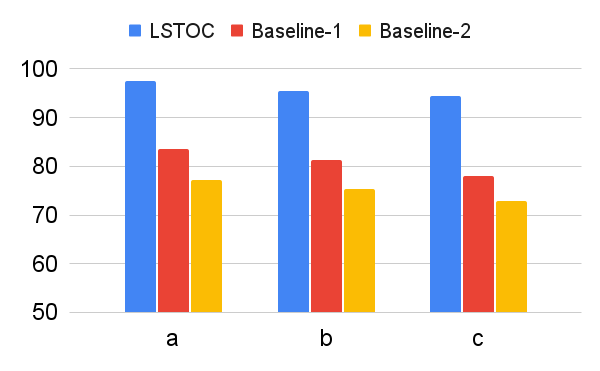}
        \label{fig:small1}
    }
    \subfigure[Office Task 2]{
        \centering
        \includegraphics[width=1.5in]{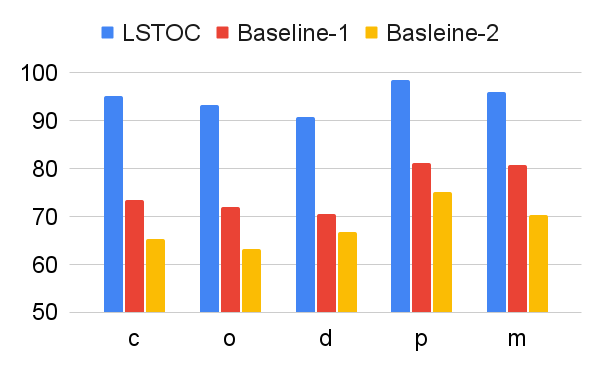}
        \label{fig:small2}
    }
    
    \subfigure[Crafter Task 1]{
        \centering
        \includegraphics[width=1.5in]{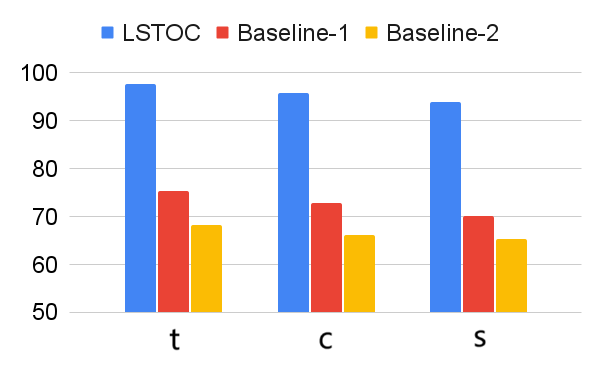}
        \label{fig:small3}
    }
    \subfigure[Crafter Task 3]{
        \centering
        \includegraphics[width=1.5in]{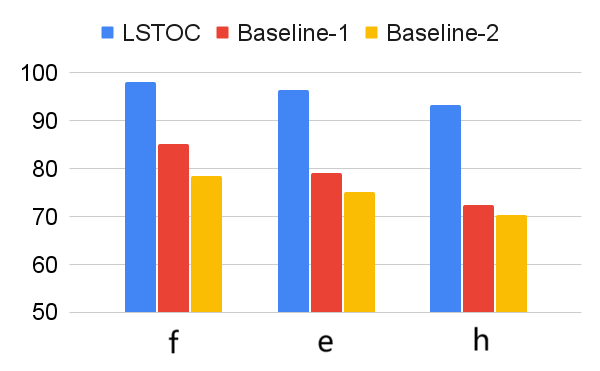}
        \label{fig:small4}
    }
    \vspace{-10pt}
    \caption{Comparison on accuracy of subgoal learning.}
    \label{fig:small_contrastive_acc}
    \vspace{-10pt}
\end{figure}

The comparison on subgoal learning is shown in Figure \ref{fig:small_contrastive_acc}, and the results for all the tasks are shown in Figure \ref{fig:contrastive_acc} in Appendix. In every figure, LSTOC and baselines are compared based on the same amount of transition samples collected from the environment, which are $10^6$ samples in Letter domain, $2\times10^6$ samples in Office domain and $2.5\times10^6$ in Crafter domain. We can see that the contrastive learning method in LSTOC significantly outperforms baselines, showing the effects of FR and temporal geometric sampling. FR is important since the agent only focuses on detecting the key state of next subgoal only and FR can prevent the distraction of key states of future subgoals. The temporal distances of detected key states are learned by temporal geometric sampling. Baseline 1 performs better than Baseline 2 in Office and Crafter domains, showing that temporal geometric sampling is more important to key state detection than FR, since the task can only be accomplished when subgoals are achieved following temporal ordering constraints.

\begin{figure}
    \centering
    \subfigure[LSTOC Task 1 Letter]{
        \centering
        \includegraphics[width=1.2in, height=.9in]{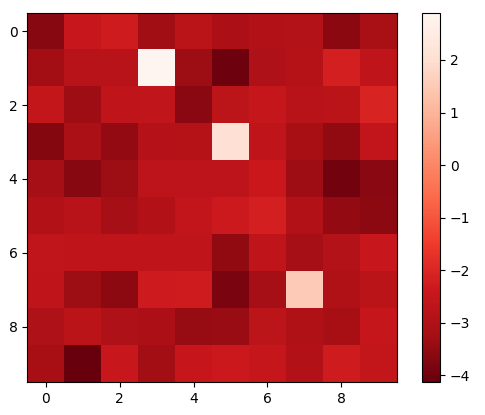}
        \label{fig:contrastive_temporal_letter_lstoc}
    }
    \hspace{15pt}
    \subfigure[Baseline Task 1 Letter]{
        \centering
        \includegraphics[width=1.2in, height=.9in]{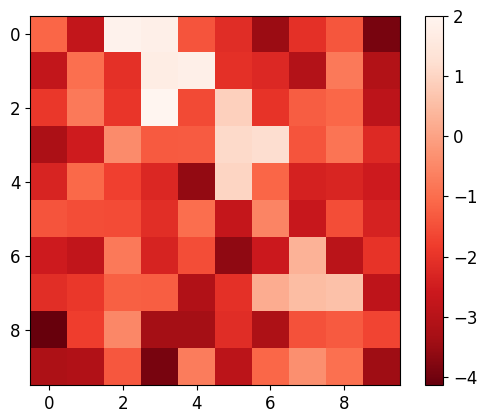}
        \label{fig:contrastive_temporal_letter_baseline}
    }
    
    \subfigure[LSTOC Task 2 Letter]{
        \centering
        \includegraphics[width=1.2in, height=.9in]{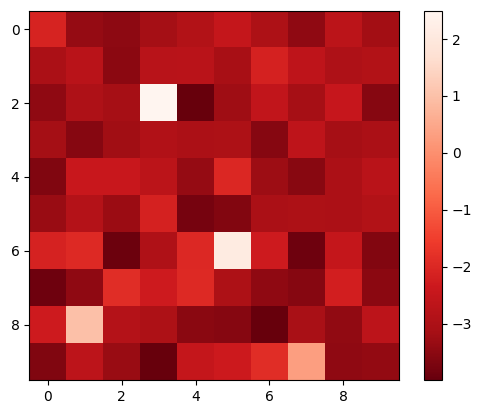}
        \label{fig:contrastive_temporal_letter2_lstoc}
    }
    \hspace{15pt}
    \subfigure[Baseline Task 2 Letter]{
        \centering
        \includegraphics[width=1.2in, height=.9in]{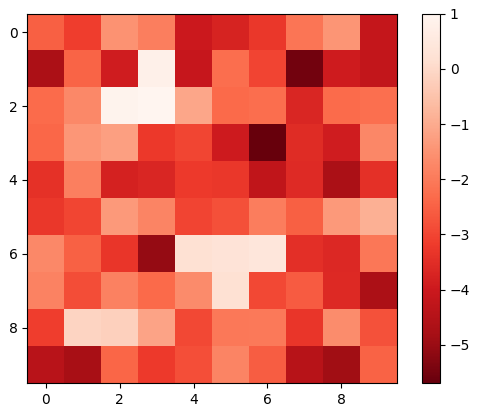}
        \label{fig:contrastive_temporal_letter2_baseline}
    }

    \subfigure[LSTOC Task 1 Office]{
        \centering
        \includegraphics[width=1.4in]{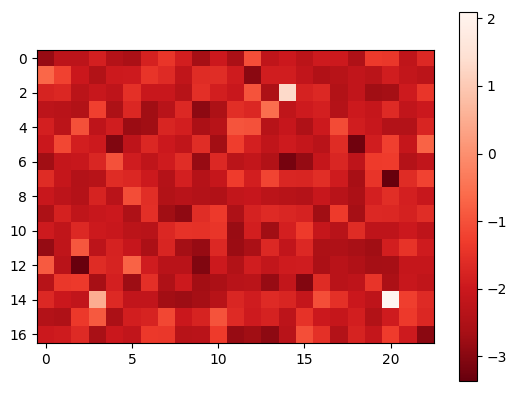}
        \label{fig:contrastive_temporal_office_lstoc}
    }
    \subfigure[Baseline Task 1 Office]{
        \centering
        \includegraphics[width=1.4in]{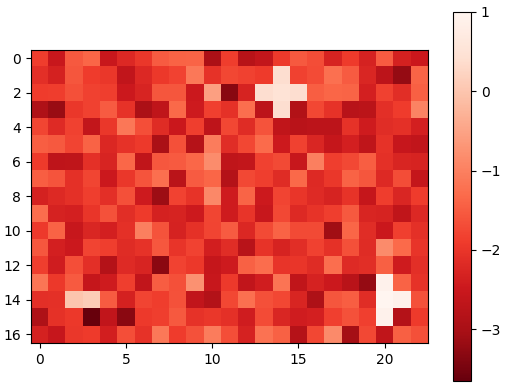}
        \label{fig:contrastive_temporal_office_baseline}
    }
    \vspace{-10pt}
    \caption{Comparison of the proposed contrastive learning method in LSTOC and baseline on learning temporal distance. The value in each grid is the output of importance function $f_{\omega}$. The higher the value is, the smaller the temporal distance is, meaning that the input state is closer to the initial state.}
    \label{fig:contrastive_temporal}
    \vspace{-10pt}
\end{figure}

\subsubsection{Visualization}
\label{sec:vis_temporal}
Furthermore, we visualize the performance of the proposed contrastive learning method in LSTOC in Figure \ref{fig:contrastive_temporal}, demonstrating its performance on detecting key states and learning their temporal distances. We use the exploration policy $\pi_{\text{exp}}$ to collect a fixed number of trajectories labeled with task completion results, based on which we apply the proposed contrastive learning method to learn key states of subgoals. The number of collected trajectories for Letter domain is $5000$ and the number in Office domain is $8000$. The proposed contrastive learning method in LSTOC is compared with the baseline which first computes the FR of trajectories and then use classical contrastive object \cite{zhang2023learning} to detect key states. Specifically, in Figure \ref{fig:contrastive_temporal}, the task in the first row is the task 1 in Letter domain in Figure \ref{fig:task_letter}, and the positions of "a", "b" and "c" are $(3, 1), (5, 2)$ and $(7, 7)$. In the second row, the task is the task 2 in Letter domain of Figure \ref{fig:task_letter}, and positions of "a", "b", "c" and "d" are $(3, 2), (1, 8), (7, 9)$ and $(5, 6)$. In the third row, the task is the task 1 in Office domain of Figure \ref{fig:task_office}, and positions of "c", "o" and "p" are $(20, 14), (14, 2)$ and $(3, 14)$. Note that the positions of subgoals are unknown to the agent initially. The contrastive learning is conducted on a fixed set of trajectories. 

In Figure \ref{fig:contrastive_temporal}, the output of importance function $f_{\omega}$ visualized in every grid is inversely proportional to the temporal distance (ordering) of input state. In the evaluation of LSTOC, the positions of subgoals are all detected correctly. Besides, the key state of first subgoal of every task has significantly highest value at the output of $f_{\omega}$ and hence has the smallest temporal distance. The relative temporal distances of key states learned by the LSTOC obey the temporal orderings in the task specification. Obviously, we can see that compared with baseline, the proposed contrastive learning method in LSTOC can detect key states of subgoals accurately, while the baseline method is influenced by neighboring states. Besides, the output of $f_{\omega}$ trained by the proposed method can clearly reflect the temporal orderings of detected subgoals in the task specification (the higher the value of $f_{\omega}$ is, the higher the temporal ordering will be), while the baseline has some ambiguity on reflecting subgoal temporal orderings in some cases.

However, we can see that when using the proposed method in LSTOC, the key states of subgoals which have big temporal distances (far away from initial state) cannot stand out at the output of $f_{\omega}$ among with other neighboring non-important states, e.g., in Figure \ref{fig:contrastive_temporal_letter2_lstoc} and \ref{fig:contrastive_temporal_office_lstoc}. This is because the contrastive learning is conducted on a fixed set of trajectories, which can only be used to detect the first or next subgoal to achieve. That is also the reason why we propose to detect subgoals one-by-one by building a subgoal tree. 

\begin{figure}
    \centering
    \subfigure[Letter Task 1]{
        \centering
        \includegraphics[width=1.5in]{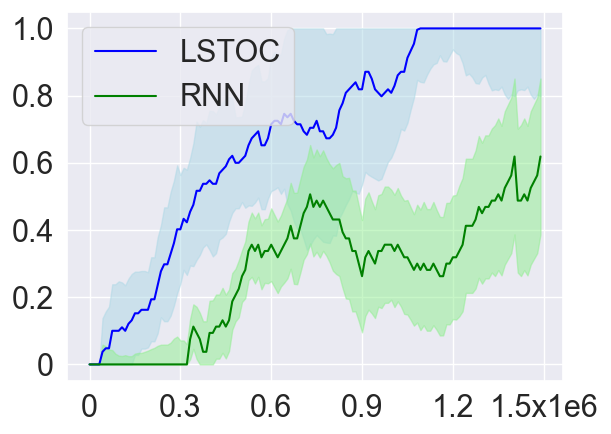}
    }
    \subfigure[Office Task 2]{
        \centering
        \includegraphics[width=1.5in]{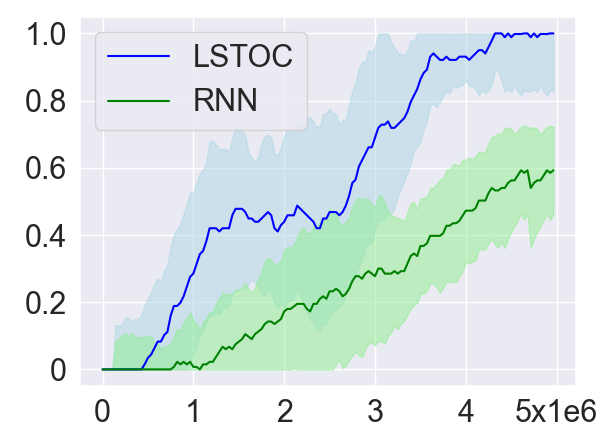}
    }
    
    \subfigure[Crafter Task 1]{
        \centering
        \includegraphics[width=1.5in]{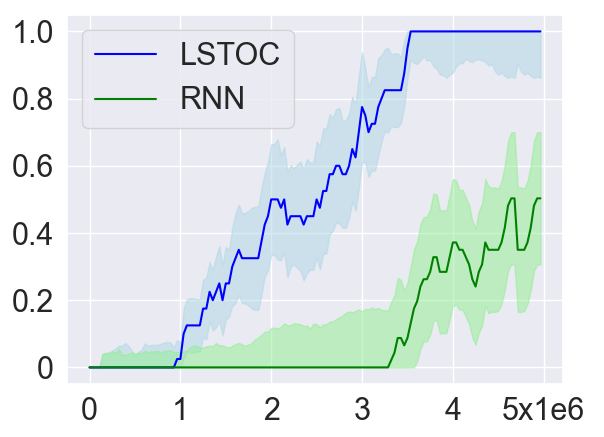}
    }
    \subfigure[Crafter Task 3]{
        \centering
        \includegraphics[width=1.5in]{plot_minihack_task3.png}
    }
    \vspace{-5pt}
    \caption{Comparison of efficiency on task solving. The y-axis is success rate of completing the task. The x-axis is the environmental step.}
    \label{fig:small_learning_efficiency}
    \vspace{-10pt}
\end{figure}

\subsubsection{Learning Efficiency}
\label{sec:learning_efficiency}
As introduced in Section \ref{sec:exp_policy}, in order to collect more trajectories conditioned on the working node, the LSTOC adds auxiliary reward to every visit to a discovered key state which can make the agent achieve the working node in $\mathcal{T}_{\varphi}$. This can significantly accelerate the agent's learning to solve the task composed by temporally extended subgoals. In order to verify this, we compare LSTOC with a baseline method which uses an RNN-based policy trained by PPO algorithm to solve the task. The results for selected tasks are shown in Figure \ref{fig:small_learning_efficiency}, with complete results shown in Figure \ref{fig:learning_efficiency} in Appendix. We can see that the LSTOC can solve every task significantly faster than the baseline. This is because the baseline does not specifically learn the subgoal and does not use auxiliary reward to resolve the issue of reward sparsity. 

In addition, we also evaluate the average number of trajectories (both positive and negative) collected until every subgoal is learned (i.e., the condition for executing Line 11 of Algorithm \ref{alg:lstoc} is met). The baseline method uses a fixed exploration policy $\bar{\pi}_{\text{exp}}$ to collect trajectories, based on which the contrastive learning in Section \ref{sec:discover} is applied to learn subgoals. Specifically, the policy $\bar{\pi}_{\text{exp}}$ is only trained by the reward of task completion before the first tree expansion is conducted. The comparison results are shown in Table \ref{tab:subgoal_efficiency_traj}. Compared with LSTOC, the difference of the baseline is that exploration policy $\bar{\pi}_{\text{exp}}$ is not further tuned to collect more trajectories conditioned on the working node. In this case, to discover subgoals along the branch of the task FSM which is harder to achieve can cost more trajectory data. This explains the advantage of LSTOC in Table \ref{tab:subgoal_efficiency_traj}.

\begin{table}
    \centering
    \begin{tabular}{l|c|c}
                    & LSTOC & Baseline
    \\\hline
     Task 2 Letter  & $18123\pm2915$ & $39256\pm2567$ \\
     Task 2 Office  & $32195\pm5623$ & $56232\pm4621$ \\
     Task 3 Crafter & $33562\pm4482$ & $50172\pm5215$
    \end{tabular}
    \caption{Comparison of Subgoal Learning Efficiency}
    \label{tab:subgoal_efficiency_traj}
\end{table}

\section{Correctness and Limitation}
\label{sec:correctness_and_limitation}
Empirically, as long as $N_T$ is large enough and randomness of exploration policy $\pi_{\text{exp}}$ is sufficient, the contrastive learning in \eqref{con-obj} can always discover the correct key state of next subgoal at every working node. Then, the correct subgoal tree can be built and labeling function can be correctly obtained by solving the ILP problem. Specifically, the randomness of $\pi_{\text{exp}}$ can be guaranteed by using $\epsilon$-greedy into action selection, where $\epsilon=0.5$ is enough for all the environments. 

However, LSTOC framework still has limitations. In some cases, the labeling component can not distinguish environmental bottleneck states from hidden subgoals. In some other cases, the labeling component cannot tell the differences of symmetric branches in the given FSM. Furthermore, the trajectory collection can be problematic in some hard-exploration environments. The details of correctness and limitation are presented in Appendix \ref{sec:app_correctness_and_limitation}.

\section{Conclusion}
\label{sec:conclusion}
In this work, we propose a framework for learning hidden subgoals under temporal ordering constraints, including a new contrastive learning method and a sample-efficient learning strategy for temporally extended hidden subgoals. In the future, we will resolve the issues of extending this work into hard-exploration environments, improving its sample efficiency in environments with large state space.

\newpage
\bibliography{main}

\begin{thebibliography}{30}
\providecommand{\natexlab}[1]{#1}

\bibitem[{Baier and Katoen(2008)}]{baier2008principles}
Baier, C.; and Katoen, J.-P. 2008.
\newblock \emph{Principles of model checking}.
\newblock MIT press.

\bibitem[{Bozkurt et~al.(2020)Bozkurt, Wang, Zavlanos, and Pajic}]{bozkurt2020control}
Bozkurt, A.~K.; Wang, Y.; Zavlanos, M.~M.; and Pajic, M. 2020.
\newblock Control synthesis from linear temporal logic specifications using model-free reinforcement learning.
\newblock In \emph{2020 IEEE International Conference on Robotics and Automation (ICRA)}, 10349--10355. IEEE.

\bibitem[{Camacho et~al.(2019)Camacho, Icarte, Klassen, Valenzano, and McIlraith}]{camacho2019ltl}
Camacho, A.; Icarte, R.~T.; Klassen, T.~Q.; Valenzano, R.~A.; and McIlraith, S.~A. 2019.
\newblock LTL and Beyond: Formal Languages for Reward Function Specification in Reinforcement Learning.
\newblock In \emph{IJCAI}, volume~19, 6065--6073.

\bibitem[{Casper et~al.(2023)Casper, Davies, Shi, Gilbert, Scheurer, Rando, Freedman, Korbak, Lindner, Freire et~al.}]{casper2023open}
Casper, S.; Davies, X.; Shi, C.; Gilbert, T.~K.; Scheurer, J.; Rando, J.; Freedman, R.; Korbak, T.; Lindner, D.; Freire, P.; et~al. 2023.
\newblock Open problems and fundamental limitations of reinforcement learning from human feedback.
\newblock \emph{arXiv preprint arXiv:2307.15217}.

\bibitem[{De~Giacomo et~al.(2019)De~Giacomo, Iocchi, Favorito, and Patrizi}]{de2019foundations}
De~Giacomo, G.; Iocchi, L.; Favorito, M.; and Patrizi, F. 2019.
\newblock Foundations for restraining bolts: Reinforcement learning with LTLf/LDLf restraining specifications.
\newblock In \emph{Proceedings of the international conference on automated planning and scheduling}, volume~29, 128--136.

\bibitem[{De~Giacomo and Vardi(2013)}]{de2013linear}
De~Giacomo, G.; and Vardi, M.~Y. 2013.
\newblock Linear temporal logic and linear dynamic logic on finite traces.
\newblock In \emph{IJCAI'13 Proceedings of the Twenty-Third international joint conference on Artificial Intelligence}, 854--860. Association for Computing Machinery.

\bibitem[{Gaon and Brafman(2020)}]{gaon2020reinforcement}
Gaon, M.; and Brafman, R. 2020.
\newblock Reinforcement learning with non-markovian rewards.
\newblock In \emph{Proceedings of the AAAI conference on artificial intelligence}, volume~34, 3980--3987.

\bibitem[{Gonzalez-Aguirre et~al.(2021)Gonzalez-Aguirre, Osorio-Oliveros, Rodriguez-Hernandez, Liz{\'a}rraga-Iturralde, Morales~Menendez, Ramirez-Mendoza, Ramirez-Moreno, and Lozoya-Santos}]{gonzalez2021service}
Gonzalez-Aguirre, J.~A.; Osorio-Oliveros, R.; Rodriguez-Hernandez, K.~L.; Liz{\'a}rraga-Iturralde, J.; Morales~Menendez, R.; Ramirez-Mendoza, R.~A.; Ramirez-Moreno, M.~A.; and Lozoya-Santos, J. d.~J. 2021.
\newblock Service robots: Trends and technology.
\newblock \emph{Applied Sciences}, 11(22): 10702.

\bibitem[{Goyal et~al.(2020)Goyal, Lamb, Hoffmann, Sodhani, Levine, Bengio, and Sch{\"o}lkopf}]{goyal2020recurrent}
Goyal, A.; Lamb, A.; Hoffmann, J.; Sodhani, S.; Levine, S.; Bengio, Y.; and Sch{\"o}lkopf, B. 2020.
\newblock Recurrent Independent Mechanisms.
\newblock In \emph{International Conference on Learning Representations}.

\bibitem[{GurobiOptimization(2023)}]{gurobi2023}
GurobiOptimization, L. 2023.
\newblock Gurobi Optimizer Reference Manual.

\bibitem[{Hafner(2021)}]{hafner2021crafter}
Hafner, D. 2021.
\newblock Benchmarking the Spectrum of Agent Capabilities.
\newblock \emph{arXiv preprint arXiv:2109.06780}.

\bibitem[{Hasanbeig et~al.(2019)Hasanbeig, Kantaros, Abate, Kroening, Pappas, and Lee}]{hasanbeig2019reinforcement}
Hasanbeig, M.; Kantaros, Y.; Abate, A.; Kroening, D.; Pappas, G.~J.; and Lee, I. 2019.
\newblock Reinforcement learning for temporal logic control synthesis with probabilistic satisfaction guarantees.
\newblock In \emph{2019 IEEE 58th conference on decision and control (CDC)}, 5338--5343. IEEE.

\bibitem[{Hatanaka, Yamashina, and Matsubara(2023)}]{hatanaka2023reinforcement}
Hatanaka, W.; Yamashina, R.; and Matsubara, T. 2023.
\newblock Reinforcement Learning of Action and Query Policies with LTL Instructions under Uncertain Event Detector.
\newblock \emph{IEEE Robotics and Automation Letters}.

\bibitem[{Icarte et~al.(2018)Icarte, Klassen, Valenzano, and McIlraith}]{icarte2018using}
Icarte, R.~T.; Klassen, T.; Valenzano, R.; and McIlraith, S. 2018.
\newblock Using reward machines for high-level task specification and decomposition in reinforcement learning.
\newblock In \emph{International Conference on Machine Learning}, 2107--2116. PMLR.

\bibitem[{Li et~al.(2022)Li, Chen, Vaezipoor, Klassen, Icarte, and McIlraith}]{li2022noisy}
Li, A.~C.; Chen, Z.; Vaezipoor, P.; Klassen, T.~Q.; Icarte, R.~T.; and McIlraith, S.~A. 2022.
\newblock Noisy Symbolic Abstractions for Deep RL: A case study with Reward Machines.
\newblock In \emph{Deep Reinforcement Learning Workshop NeurIPS 2022}.

\bibitem[{Liang et~al.(2022)Liang, Shu, Lee, and Abbeel}]{liang2022reward}
Liang, X.; Shu, K.; Lee, K.; and Abbeel, P. 2022.
\newblock Reward Uncertainty for Exploration in Preference-based Reinforcement Learning.
\newblock In \emph{10th International Conference on Learning Representations, ICLR 2022}. International Conference on Learning Representations.

\bibitem[{Littman et~al.(2017)Littman, Topcu, Fu, Isbell, Wen, and MacGlashan}]{littman2017environment}
Littman, M.~L.; Topcu, U.; Fu, J.; Isbell, C.; Wen, M.; and MacGlashan, J. 2017.
\newblock Environment-independent task specifications via GLTL.
\newblock \emph{arXiv preprint arXiv:1704.04341}.

\bibitem[{Luo et~al.(2023)Luo, Mao, Wu, Lozano-P{\'e}rez, Tenenbaum, and Kaelbling}]{luo2023learning}
Luo, Z.; Mao, J.; Wu, J.; Lozano-P{\'e}rez, T.; Tenenbaum, J.~B.; and Kaelbling, L.~P. 2023.
\newblock Learning rational subgoals from demonstrations and instructions.
\newblock In \emph{Proceedings of the AAAI Conference on Artificial Intelligence}, volume~37, 12068--12078.

\bibitem[{Mnih et~al.(2015)Mnih, Kavukcuoglu, Silver, Rusu, Veness, Bellemare, Graves, Riedmiller, Fidjeland, Ostrovski et~al.}]{mnih2015human}
Mnih, V.; Kavukcuoglu, K.; Silver, D.; Rusu, A.~A.; Veness, J.; Bellemare, M.~G.; Graves, A.; Riedmiller, M.; Fidjeland, A.~K.; Ostrovski, G.; et~al. 2015.
\newblock Human-level control through deep reinforcement learning.
\newblock \emph{nature}, 518(7540): 529--533.

\bibitem[{Moskovitz, Wilson, and Sahani(2021)}]{moskovitz2021first}
Moskovitz, T.; Wilson, S.~R.; and Sahani, M. 2021.
\newblock A First-Occupancy Representation for Reinforcement Learning.
\newblock In \emph{International Conference on Learning Representations}.

\bibitem[{Ni et~al.(2021)Ni, Eysenbach, Levine, and Salakhutdinov}]{ni2021recurrent}
Ni, T.; Eysenbach, B.; Levine, S.; and Salakhutdinov, R. 2021.
\newblock Recurrent model-free rl is a strong baseline for many pomdps.

\bibitem[{Oord, Li, and Vinyals(2018)}]{oord2018representation}
Oord, A. v.~d.; Li, Y.; and Vinyals, O. 2018.
\newblock Representation learning with contrastive predictive coding.
\newblock \emph{arXiv preprint arXiv:1807.03748}.

\bibitem[{Park et~al.(2022)Park, Seo, Shin, Lee, Abbeel, and Lee}]{park2022surf}
Park, J.; Seo, Y.; Shin, J.; Lee, H.; Abbeel, P.; and Lee, K. 2022.
\newblock SURF: Semi-supervised Reward Learning with Data Augmentation for Feedback-efficient Preference-based Reinforcement Learning.
\newblock In \emph{International Conference on Learning Representations}.

\bibitem[{Ronca et~al.(2022)Ronca, Paludo~Licks, De~Giacomo et~al.}]{ronca2022markov}
Ronca, A.; Paludo~Licks, G.; De~Giacomo, G.; et~al. 2022.
\newblock Markov Abstractions for PAC Reinforcement Learning in Non-Markov Decision Processes.
\newblock In \emph{Proceedings of the Thirty-First International Joint Conference on Artificial Intelligence, IJCAI 2022}, 3408--3415. International Joint Conferences on Artificial Intelligence.

\bibitem[{Schulman et~al.(2017)Schulman, Wolski, Dhariwal, Radford, and Klimov}]{schulman2017proximal}
Schulman, J.; Wolski, F.; Dhariwal, P.; Radford, A.; and Klimov, O. 2017.
\newblock Proximal policy optimization algorithms.
\newblock \emph{arXiv preprint arXiv:1707.06347}.

\bibitem[{Sun et~al.(2023)Sun, Yang, Jiralerspong, Malenfant, Alsbury-Nealy, Bengio, and Richards}]{sun2023contrastive}
Sun, C.; Yang, W.; Jiralerspong, T.; Malenfant, D.; Alsbury-Nealy, B.; Bengio, Y.; and Richards, B.~A. 2023.
\newblock Contrastive Retrospection: honing in on critical steps for rapid learning and generalization in RL.
\newblock In \emph{Thirty-seventh Conference on Neural Information Processing Systems}.

\bibitem[{Sutton and Barto(2018)}]{sutton2018reinforcement}
Sutton, R.~S.; and Barto, A.~G. 2018.
\newblock \emph{Reinforcement learning: An introduction}.
\newblock MIT press.

\bibitem[{Umili, Capobianco, and De~Giacomo(2023)}]{umili2023grounding}
Umili, E.; Capobianco, R.; and De~Giacomo, G. 2023.
\newblock Grounding LTLf specifications in image sequences.
\newblock In \emph{Proceedings of the International Conference on Principles of Knowledge Representation and Reasoning}, volume~19, 668--678.

\bibitem[{Xu et~al.(2021)Xu, Wu, Ojha, Neider, and Topcu}]{xu2021active}
Xu, Z.; Wu, B.; Ojha, A.; Neider, D.; and Topcu, U. 2021.
\newblock Active finite reward automaton inference and reinforcement learning using queries and counterexamples.
\newblock In \emph{Machine Learning and Knowledge Extraction: 5th IFIP TC 5, TC 12, WG 8.4, WG 8.9, WG 12.9 International Cross-Domain Conference, CD-MAKE 2021, Virtual Event, August 17--20, 2021, Proceedings 5}, 115--135. Springer.

\bibitem[{Zhang and Kashima(2023)}]{zhang2023learning}
Zhang, G.; and Kashima, H. 2023.
\newblock Learning state importance for preference-based reinforcement learning.
\newblock \emph{Machine Learning}, 1--17.

\end{thebibliography}

\newpage
\appendix
\onecolumn

\section{Labeling Component of LSTOC}
\label{sec:app_labeling}
The FSM $\mathcal{M}_{\varphi}$ is the FSM specification of subgoal temporal dependencies given by the user, where each node is described by a subgoal semantic symbol in $\mathcal{G}$. Denote the state space of $\mathcal{M}_{\varphi}$ as $\mathcal{U}$ which has $U$ states. The transition function of $\mathcal{M}_{\varphi}$ is defined as $\delta_{\varphi}:\mathcal{U}\times\mathcal{G}\times\mathcal{U}\to\{0,1\}$, and the transition from state $u$ to $u'$ conditioned on symbol $g$ is expressed as $\delta_{\varphi}(u, g, u')=1$, where $g$ is the index of symbol in $\mathcal{G}$. 

Assume that the set of discovered satisfying sequences ${\mathcal{P}}_{S}$ has $P$ sequences, and the $m$-th sequence has $l_m$ elements. Note that $\mathcal{P}_S$ consists of all the paths in the subgoal tree $\mathcal{T}_{\varphi}$ which is built by the subgoal learning component of LSTOC framework. The set $\hat{\mathcal{S}}_K$ contains the discovered key states of subgoals. The target of labeling component of LSTOC is to determine the mapping from $\hat{\mathcal{S}}_K$ to $\mathcal{G}$, making every sequence in $\mathcal{P}_S$ lead to an accepting state in $\mathcal{M}_{\varphi}$ and hence yielding the labeling function $L_{\varphi}$.

Now we start formulating the integer linear programming (ILP) problem for learning the labeling function. The binary variables of this ILP problem are composed by state transition variables $u_{m,n,i,j}$ and mapping variables $v_{k,l}$, where $u_{m,n,i,j}=1$ denotes that the $n$-th element of $m$-th sequence of $\mathcal{P}_S$ makes the agent transit from state $i$ to $j$ over FSM ${\mathcal{M}}_{\varphi}$, and $v_{k,l}=1$ denotes that $k$-th state in $\hat{\mathcal{S}}_K$ is mapped to $l$-th symbol in $\mathcal{G}$. Based on their definitions, we can first have 5 constraints on these binary variables:
\begin{eqnarray}
    &&\sum_{i,j=1}^{U}u_{m,n,i,j}=1, \hspace{5pt} \forall m=1,\ldots,|\mathcal{P}_S|, n=1,\ldots,l_n \label{label0} \\
    &&\sum_{j=1}^{U}u_{m,1,1,j}=1, \hspace{5pt} \forall m=1,\ldots,|\mathcal{P}_S| \label{label1} \\
    &&\sum_{i=1}^U u_{m,n,i,j}=\sum_{i=1}^U u_{m,n+1,j,i}, \hspace{5pt} \forall m=1,\ldots,|\mathcal{P}_S|, n=1,\ldots,l_n-1, j=1,\ldots,U \label{label2} \\
    &&\sum_{l=1}^{|\mathcal{G}|} v_{k,l}\le 1, \hspace{5pt} \forall k=1,\ldots,|\hat{\mathcal{S}}_K| \label{label3} \\
    &&\sum_{k=1}^{|\hat{\mathcal{S}}_K|} v_{k,l}=1, \hspace{5pt} \forall l=1,\ldots,|\mathcal{G}| \label{label4}
\end{eqnarray}
where these constraints mean: 1) every element of every sequence in $|\mathcal{P}_S|$ makes a transition, including staying at the same state of FSM ${\mathcal{M}}_{\varphi}$; 2) the first element of every sequence is in the first state of ${\mathcal{M}}_{\varphi}$; 3) for any pair of consecutive elements of every sequence, the out-going state of the previous element is the same as the in-coming state of the other one; 4) every state in $\hat{\mathcal{S}}_K$ is mapped to at most one semantic symbol in $\mathcal{G}$, since some discovered key states may be redundant; 5) every symbol in $\mathcal{G}$ is associated with one discovered key state in $\hat{\mathcal{S}}_K$. 

Since the transition variables $u_{m,n,i,j}$ and mapping variables $v_{k,l}$ must be consistent with the state transitions of ${\mathcal{M}}_{\varphi}$ ($\delta_{\varphi}$), we have another set of constraints:
\begin{equation}
    u_{m,n,i,j}\le\delta_{\varphi}(i,l,j)\cdot v_{k,l} \label{label5}
\end{equation}
where $m,n,i,j$ and $k,l$ have the same range of values as above constraints.

Finally, we have another set of constraints which make sure that the last element of every sequence in $\mathcal{P}_S$ makes the agent stay in any accepting state of FSM $\mathcal{M}_{\varphi}$. Then, we have
\begin{equation}
    \sum_{j\in\mathcal{U}_F} u_{m,n,i,j} = 1 \label{label6}
\end{equation}
where $\mathcal{U}_F$ denotes the set of accepting states of $\mathcal{M}_{\varphi}$. In order to ignore the states in $\hat{\mathcal{S}}_K$ not associated with any subgoals during mapping, such as bottleneck state in the environmental layout, we use the sum of mapping variables as the objective:
\begin{equation}
    \sum_{k=1}^{|\hat{\mathcal{S}}_K|}\sum_{l=1}^{|\mathcal{G}|} v_{k,l} \label{label7}
\end{equation}
The formulated ILP problem has the objective \eqref{label7} and constraints \eqref{label0}-\eqref{label6}. We solve it by Gurobi solver \citep{gurobi2023}.

\section{Algorithms}
\label{sec:app_algo}
We present the algorithm tables of the proposed framework in this section. Since the agent is not familiar with the environment in advance and the optimal selection of $N_T$ is not clear, we design a main loop to try to call LSTOC (described in Algorithm \ref{alg:lstoc}) with different values of $N_T$ increasing incrementally. In implementation, we set $N_T=80$ and $H_T=20$ for every domain.

The subgoal discovery process at a working node is described in Algorithm \ref{alg:contrastive}. It is same as the process described in Section \ref{sec:discover}. The inputs $\mathcal{D}_P$ and $\mathcal{D}_N$ only contain trajectories conditioned on the current working node. In line 3, the discriminative state representation is first pre-trained based on data in $\mathcal{B}_P\cup\mathcal{B}_N$. For line 5 and 6, In the process of FR, trajectories in $\mathcal{D}$ are processed to only keep the part after achieving the working node (the red part in Figure \ref{fig:traj_exp} as an example), then the FR of every processed trajectory is first computed by removing repetitive states. Then, from line 7 to 13, the importance function $f_{\omega}$ is iteratively trained by the objective formulated in \eqref{con-obj}. In implementation, the number of iterations ($L$) is set to be $700$ for every domain. The batch size of negative samples ($B$) is chosen to be $64$. Due to the simple architecture of $f_{\omega}$, the learning rate of each iteration is set to be $0.01$. In line 14, the state with highest value of $f_{\omega}$ (highest temporal ordering) is selected as the discovered key state of next subgoal and returned to LSTOC in line 15.

The LSTOC is described in Algorithm \ref{alg:lstoc}. In line 5, the agent first collects sufficient number of trajectories from the environment. The loop starting at line 6 is the iterating process of building subgoal tree $\mathcal{T}_{\varphi}$ which will terminate when 1) every positive trajectory is explained by $\mathcal{T}_{\varphi}$ (line 8) and ILP problem for symbol grounding is successfully solved (line 9); or 2) $\mathcal{T}_{\varphi}$ is wrongly built (line 37) with condition (***) introduced in the following paragraph. The loop starting at line 7 is the loop of collecting sufficient number of positive trajectory for subgoal discovery at the current working node $v_w$. From line 14 to 16, trajectories conditioned on $v_w$ are selected from $\mathcal{B}_P$ and $\mathcal{B}_N$ with explained trajectories discarded, and stored as $\mathcal{D}_P$ and $\mathcal{D}_N$. In line 17, if every trajectory in $\mathcal{D}_P$ which follows the path of $v_w$ can lead to the accomplishment of the task (getting a positive label), the path $p_w$ will be a newly discovered satisfying path and tree expansion will be moved back to the parent node of $v_w$, initiating another iteration. In line 21, if the condition (**) of being fully explored is met, the tree expansion will be moved to the parent node of $v_w$ and starts another iteration of expansion. In line 24, if the condition of subgoal discovery at $v_w$ is not met, more trajectories will be collected by Explore process and exploration policy will be further trained to make more trajectories conditioned on $v_w$ to be collected. Otherwise, the subgoal discovery process will be called in line 32. After new key state of subgoal is discovered, from line 33 to 39, $\mathcal{T}_{\varphi}$ and set of key states $\hat{\mathcal{S}}_K$ will be expanded by adding a new node $v_{\text{new}}$. Then, in line 40 and 41, the working node is moved to $v_{\text{new}}$.

The condition (*) in Algorithm \ref{alg:lstoc} states that $p_w$ is a newly discovered satisfying sequence (concept defined in Section \ref{sec:tl}) and no further expansion at the current working node is needed. The condition (**) in Algorithm \ref{alg:lstoc} states that every positive trajectory conditioned on $v_w$ can be explained and no further expansion at the current working node is needed. The condition (***) in Algorithm \ref{alg:lstoc} refers to the situations where $\mathcal{T}_{\varphi}$ is wrong built. The condition (***) becomes true when the longest path in $\mathcal{T}_{\varphi}$ is longer than that in $\mathcal{M}_{\varphi}$ or the largest degree of nodes in $\mathcal{T}_{\varphi}$ is larger than that in $\mathcal{M}_{\varphi}$.

\begin{algorithm}
\caption{Main Loop}
\label{alg:main}
\begin{algorithmic}[1]
\State {\bf Require} $M_{\varphi}$: FSM representing the temporal dependencies of subgoals; $K_T$: threshold increment of each round of LSTOC;
\State $N_T\leftarrow K_T$;
\While{True}
\State result, dict $\leftarrow$ LSTOC$(N_T, M_{\varphi})$; {\it \% Call Algorithm \ref{alg:lstoc}}
\If{result is True}
\State {\bf Return} dict;  {\it \% Hidden subgoals and their semantic symbols are learned and returned}
\Else
\State $N_T\leftarrow N_T + K_T$; {\it \% threshold $N_T$ is increased to start another call of LSTOC algorithm}
\EndIf
\EndWhile
\end{algorithmic}
\end{algorithm}

\begin{small}
\begin{algorithm}
\caption{Contrastive$(\mathcal{D}_P, \mathcal{P}_N, v_w)$}
\label{alg:contrastive}
\begin{algorithmic}[1]
\State {\bf Input} $\mathcal{D}_P (\mathcal{D}_N)$: buffer of selected positive (negative) trajectories for subgoal discovery; $v_w$: current working node; 
\State {\bf Notation} FR(): function for computing FR introduced in Section \ref{sec:discover}; $o_{\theta}$: representation for non-symbolic state or observation; $f_{\omega}$: function inversely proportional to the temporal distance; $B$: number of negative states for formulating the contrastive objective; $\gamma$: discount factor of the working environment;
\State Pre-train the state representation $o_{\theta}$ based on InfoNCE loss \citep{oord2018representation};
\State Initialize $f_{\omega}$;
\State $\tilde{\mathcal{D}}_P\leftarrow\text{FR}(\mathcal{D}_P, v_w)$; {\it \% Process introduced in the "FR" paragraph of Section \ref{sec:discover}}
\State $\tilde{\mathcal{D}}_N\leftarrow\text{FR}(\mathcal{D}_N, v_w)$; \For{$i=1,2,\ldots, L$}
\State Sample $\tau^+\sim\tilde{\mathcal{D}}_P$;
\State $\{s_{t^+}\}\leftarrow$ sample one state from $\tau^+$ with $t^+\sim\text{Geom}(1-\gamma)$;
\State $\{s_{t_i^-}\}_{i=1}^B\leftarrow$ sample $B$ trajectories from $\tilde{\mathcal{D}}_N$ and sample $B$ states from these trajectories with $t_i^-\sim\text{Geom}(1-\gamma)$;
\State Formulate objective \eqref{con-obj} with $\{s_{t^+}\}$ and $\{s_{t_i^-}\}_{i=1}^B$;
\State Compute the gradient of \eqref{con-obj} to train $f_{\omega}$;
\EndFor
\State $\hat{s}_{\text{new}}\leftarrow$ select the state with highest output of $f_{\omega}$ among trajectories in $\mathcal{D}_P$;
\State {\bf Return} $\hat{s}_{\text{new}}$
\end{algorithmic}
\end{algorithm}
\end{small}

\begin{algorithm}
\caption{LSTOC$(N_T, M_{\varphi})$}
\label{alg:lstoc}
\begin{algorithmic}[1]
\State {\bf Input} $N_T$: the threshold of number of unexplained positive trajectories to initiate a subgoal discovery; $M_{\varphi}$: the FSM for temporal dependencies of subgoals;
\State {\bf Notation} $\mathcal{T}_{\varphi}$: subgoal tree; $v_w$: working node on $\mathcal{T}_{\varphi}$; $p_w$: working path (set of discovered key states along the path from $v_0$ to $v_w$ on $\mathcal{T}_{\varphi}$); $\mathcal{P}_S$: set of discovered satisfying sequences; $\mathcal{D}_P (\mathcal{D}_N)$: selected unexplained trajectories for subgoal discovery; Explore$(X)$: collect trajectories from the environment with policy $\pi_{\text{exp}}$ until $X$ positive trajectories are collected; $B_T$: number of positive trajectories collected in each exploration; other notations are introduced in the caption of Figure \ref{fig:ground_diagram};
\State Initialize $\mathcal{T}_{\varphi}$ with root node $v_0$ and set $v_w\leftarrow v_0$;
\State Initialize $p_w\leftarrow [], \hat{\mathcal{S}}_K\leftarrow\{\}$ and $\mathcal{P}_S\leftarrow \{\}$;
\State $\mathcal{B}_P, \mathcal{B}_N\leftarrow$ Explore$(N_T)$; 
\While{True}
\While{True}
\If{$v_w == v_0$ and $\mathcal{B}_P$ can all be explained by $\mathcal{P}_S$}
\State $\text{result}, L_{\varphi}\leftarrow$ ILP$(M_{\varphi}, \mathcal{P}_S)$;
\If{result is True}
\State {\bf Return} True, $\{L_{\varphi}, \hat{\mathcal{S}}_K\}$ {\it \% Hidden subgoals and temporal orderings are all correctly learned}
\EndIf
\EndIf
\State $\mathcal{D}_P\leftarrow$ select trajectories conditioned on $v_w$ from $\mathcal{B}_P$;
\State $\mathcal{D}_P\leftarrow$ discard trajectories in $\mathcal{D}_P$ which can be explained by paths in $\mathcal{P}_S$;
\State $\mathcal{D}_N\leftarrow$ select trajectories conditioned on $v_w$ from $\mathcal{B}_N$;
\If{$p_w$ is a satisfying sequence} {\it \% Condition (*)}
\State $\mathcal{P}_S\leftarrow\mathcal{P}_S\cup \{p_w\}$;
\State $v_w\leftarrow$ parent of $v_w$;
\State $p_w\leftarrow p_w[:-1]$; {\it \% Discard the last element of path $p_w$}
\ElsIf{$v_w$ is fully explored} {\it \% Condition (**)}
\State $v_w\leftarrow$ parent of $v_w$;
\State $p_w\leftarrow p_w[:-1]$;
\ElsIf{$|\mathcal{D}_P|< N_T$}
\State $\mathcal{B}_P', \mathcal{B}_N'\leftarrow$ Explore$(B_T)$;
\State $\mathcal{B}_P\leftarrow\mathcal{B}_P\cup\mathcal{B}_P', \mathcal{B}_N\leftarrow\mathcal{B}_N\cup\mathcal{B}_N'$
\State PPO$(\mathcal{B}_P'\cup\mathcal{B}_N)$; {\it \% Train $\pi_{\text{exp}}$ with PPO}
\Else
\State Break;
\EndIf
\EndWhile
\State $\hat{s}_{\text{new}}\leftarrow$ Contrastive$(\mathcal{D}_P, \mathcal{D}_N, v_w)$; {\it \% Call Algorithm \ref{alg:contrastive} to discover next subgoal}
\If{$\hat{s}_{\text{new}}$ not in $\hat{\mathcal{S}}_K$}
\State $\hat{\mathcal{S}}_K\leftarrow\hat{\mathcal{S}}_K\cup\{\hat{s}_{\text{new}}\}$;
\EndIf
\State Add a new node $n_{\text{new}}$ to $\mathcal{T}_{\varphi}$ as a child of $v_w$, labeled with $\hat{s}_{\text{new}}$;
\If{$\mathcal{T}_{\varphi}$ is inconsistent with $M_{\varphi}$} {\it \% Condition (***)}
\State {\bf Return} False, None  {\it \% Subgoals are wrongly learned and another round of LSTOC is needed}
\EndIf
\State $v_w\leftarrow n_{\text{new}}$;
\State $p_w\leftarrow p_w\cup[\hat{s}_{\text{new}}]$;
\EndWhile
\end{algorithmic}
\end{algorithm}

\begin{figure}
    \centering
    \subfigure[Letter]{
        \centering
        \includegraphics[width=1.in, height=1.in]{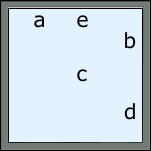}
        \label{fig:env_letter}
    }
    \hspace{5pt}
    \subfigure[Office]{
        \centering
        \includegraphics[width=2in, height=1.in]{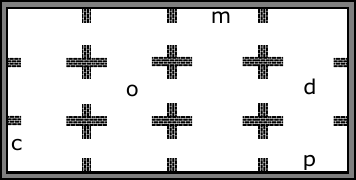}
        \label{fig:env_office}
    }
    \hspace{5pt}
    \subfigure[Crafter]{
        \centering
        \includegraphics[width=1in, height=1in]{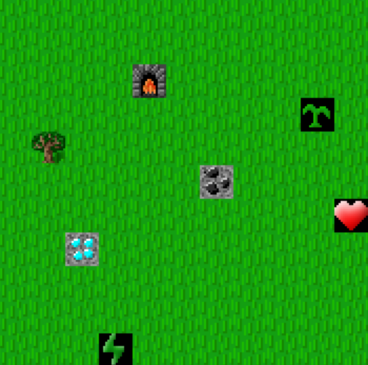}
        \label{fig:env_crafter}
    }
    \caption{Domains. The agent has full or partial observations in Letter domain, and has partial observation in Office and Crafter domain. The Crafter has pixel-based observations to the agent.}
    \label{fig:environments1}
\end{figure}

\section{Environments}
\label{sec:env}
Environment domains adopted in the experiments include both grid-based and pixel-based observations. Note that in all the environment domains, there is no labelling function which maps from agent's observation into any letter or items, so letters or items are all hidden to the agent. 

\noindent
{\bf Letter.}
The first environment domain is the {\it Letter}. As shown in Figure \ref{fig:env_letter}, there are multiple letters allocated in the map. In this domain, the observation of the agent can be the full or partial map, which is an image-based tensor without any specific information on locations of letters. If the map has the size of $m\times n$ with $k$ letters, the observation is a $m\times n\times (k+1)$ binary tensor. The agent's actions include movements in four cardinal directions. Only a subset of letters in the map is used as task subgoals, making the problem more challenging. 

\noindent
{\bf Office.}
The second environment domain is the {\it Office}. It is a variant of office game widely used in previous papers \cite{icarte2018using}. As shown in Figure \ref{fig:env_office}, in this domain, the agent only as {\it partial observation} of the environment, which is a $5\times 5$ grid centric to the agent. There are 12 rooms in the map with some segments of walls as obstacles. Five letters, "c", "o", "m", "d" and "p", represent coffee, office, mailbox, desk, and printer, which are randomly located in these rooms. One room has at most one letter. The task subgoals may use part of these five letters. The observation of the agent is a $5\times 5\times 7$ binary tensor, including letters, wall and agent. The size of whole map is $19\times 25$.

\noindent
{\bf Crafter.}
Crafter is a powerful sandbox for designing custom environments \citep{hafner2021crafter}. The agent there can navigate in the map to visit landmarks and pick up various of objects to finish the task. Our experiments the original environment to focus on the navigation part. An example of the screenshot is shown in Figure \ref{fig:env_crafter}. In our experiments, the map is a 20$\times$20 grid. The observation to the agent is a $5\times 5$ partial observation centric to the agent, where each grid of the observation is described by 16$\times$16 pixels. The objects include tree, sapling, coal, diamond, energy, furnace and health, short for "t", "s", "c", "d", "e", "f" and "h". The task formula is described in terms of letters for these objects. The agent needs to visit right objects in the right orders. 

\begin{figure}
    \centering
    \subfigure[Letter]{
        \centering
        \fontsize{8pt}{10pt}\selectfont
        \def\svgwidth{4in}
        \def\svgheight{1.2in}
        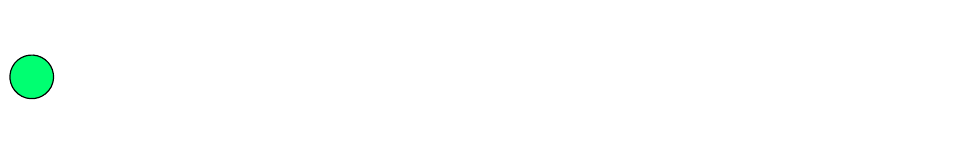
        \label{fig:task_letter}
    }

    \subfigure[Office]{
        \centering
        \fontsize{8pt}{10pt}\selectfont
        \def\svgwidth{4.in}
        \def\svgheight{1.2in}
        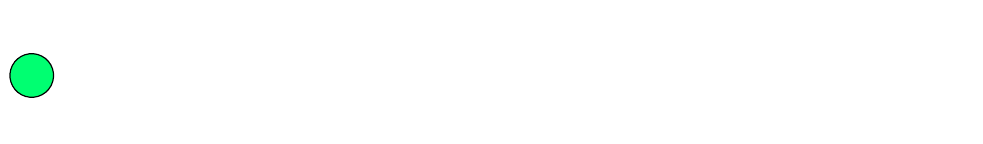
        \label{fig:task_office}
    }

    \subfigure[Crafter]{
        \centering
        \fontsize{8pt}{10pt}\selectfont
        \def\svgwidth{4.in}
        \def\svgheight{1.2in}
        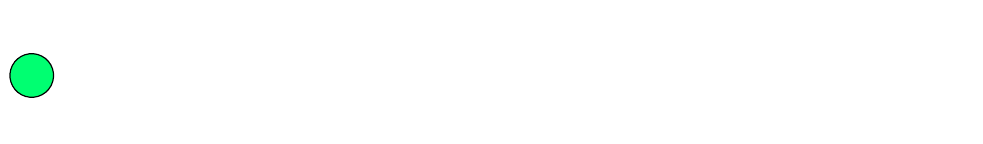
        \label{fig:task_crafter}
    }
    \caption{The FSM of subgoal temporal dependencies for the given task. The agent cannot know the position directly via the interaction with the environment.}
    \label{fig:task}
\end{figure}

\section{Tasks}
\label{sec:tasks}
The FSMs describing temporal dependencies of hidden subgoals are shown in Figure \ref{fig:task}. The LSTOC is evaluated and compared on tasks with these FSMs. The first task of every domain is always a sequential task consisting of 3 subgoals. In other tasks, there are converging and diverging branches, which are designed to evaluate the LSTOC's capability of discovering alternative branches. Some tasks, e.g., task 3 in Letter and task 2 in Office, have repetitive subgoals, which are designed to test whether LSTOC can address repetitive subgoals or not.

\section{Generalization}
\label{sec:generalization}
We evaluate the generalization capability of LSTOC in Office and Crafter domains. In the training environment, the hidden subgoals and labeling function are first learned by LSTOC framework, where Office and Crafter domains both use task 2 in Figure \ref{fig:task_office} and \ref{fig:task_crafter} to learn subgoals. Then, in the testing environment, the agent solves an unseen task with the help of auxiliary rewards given by the labeling function (learned by LSTOC in training) and task FSM. Specifically, the agent gets a positive reward whenever he visits a key state corresponding to the subgoal which can make progress toward an accepting state in the task FSM.
The testing environment is different from the training one.
In Office domain, the testing environment has different layout of wall segments compared with the training environment. In Crafter domain, new and unseen objects, including drink, sand, table and zombie, are added to the testing environment. 

\begin{figure}
    \centering
    \subfigure[Office]{
        \centering
        \includegraphics[width=1.5in]{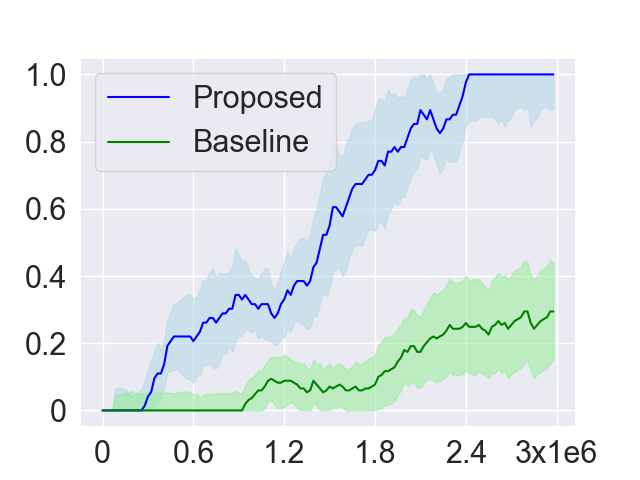}
        \label{fig:generalization_office}
    }
    \subfigure[Crafter]{
        \centering
        \includegraphics[width=1.5in]{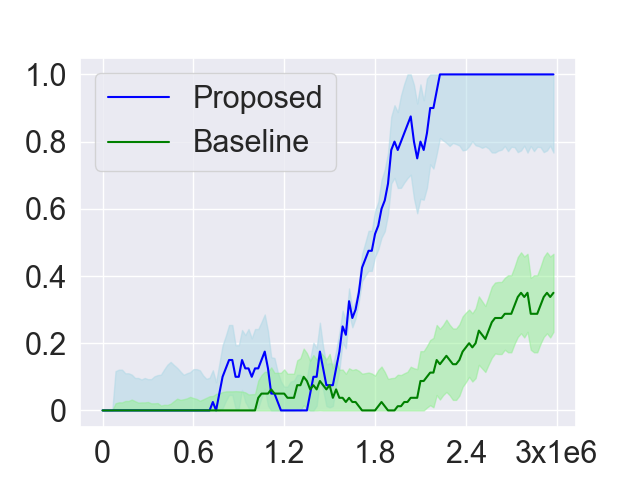}
        \label{fig:generalization_crafter}
    }
    \caption{Generalization performance. The x-axis is environmental step. The y-axis is the success rate of completing the given task unseen in the training.}
    \label{fig:generalization}
\end{figure}

In evaluation, both the proposed and baseline methods use a RNN-based agent trained by PPO algorithm. In baseline, the agent only receives a positive reward whenever the task is accomplished. However, in the proposed method, the agent additionally gets auxiliary rewards. The comparison is shown in Figure \ref{fig:generalization}. The significant acceleration of the proposed method verifies the generalization effect of subgoals learned by LSTOC framework, where the learned labeling function can adapt to changes of layout and ignore distracting unseen objects.

\begin{figure}
    \centering
    \subfigure[Task 1. Letter]{
        \centering
        \includegraphics[width=1.5in]{contrastive_letter_task1.png}
        \label{fig:contrastive_letter_task1}
    }
    \subfigure[Task 2. Letter]{
        \centering
        \includegraphics[width=1.5in]{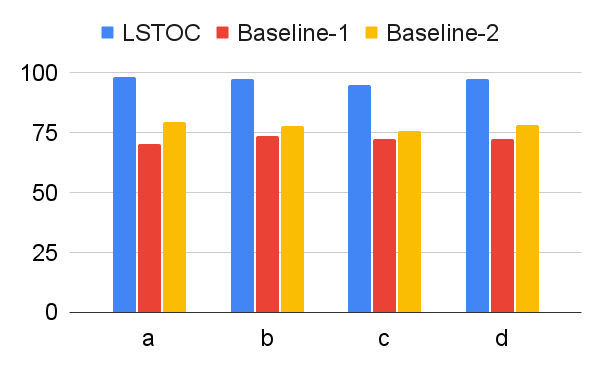}
        \label{fig:contrastive_letter_task2}
    }
    \subfigure[Task 3. Letter]{
        \centering
        \includegraphics[width=1.5in]{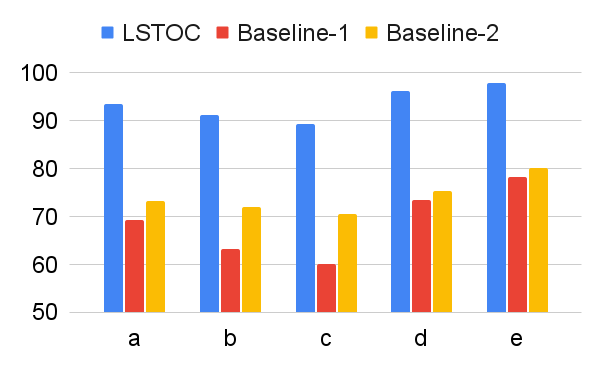}
        \label{fig:contrastive_letter_task3}
    }

    \subfigure[Task 1. Office]{
        \centering
        \includegraphics[width=1.5in]{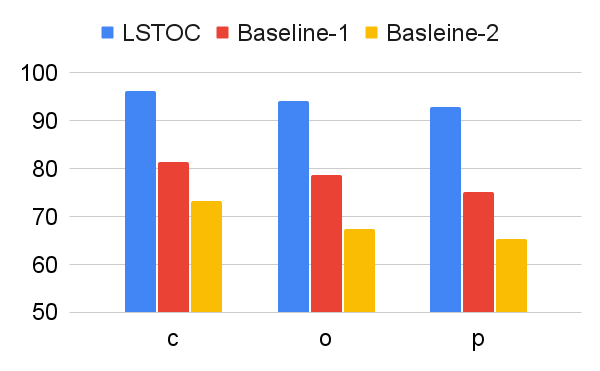}
        \label{fig:contrastive_office_task1}
    }
    \subfigure[Task 2. Office]{
        \centering
        \includegraphics[width=1.5in]{contrastive_office_task2.png}
        \label{fig:contrastive_office_task2}
    }
    \subfigure[Task 3. Office]{
        \centering
        \includegraphics[width=1.5in]{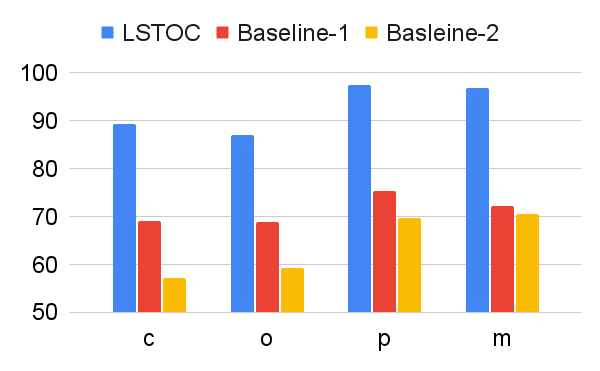}
        \label{fig:contrastive_office_task3}
    }

    \subfigure[Task 1. Crafter]{
        \centering
        \includegraphics[width=1.5in]{contrastive_crafter_task1.png}
        \label{fig:contrastive_minihack_task1}
    }
    \subfigure[Task 2. Crafter]{
        \centering
        \includegraphics[width=1.5in]{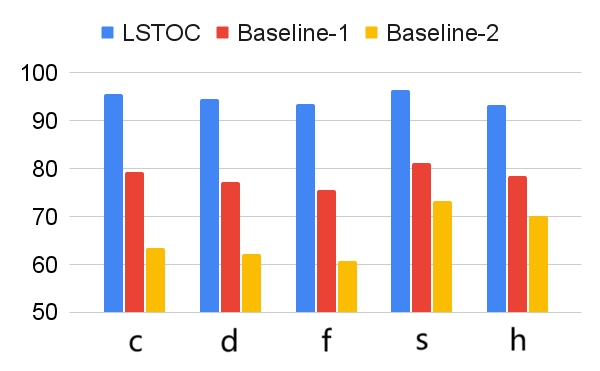}
        \label{fig:contrastive_minihack_task2}
    }
    \subfigure[Task 3. Crafter]{
        \centering
        \includegraphics[width=1.5in]{contrastive_crafter_task3.png}
        \label{fig:contrastive_minihack_task3}
    }
    \caption{Performance of contrastive learning. The vertical axis in every figure denotes the accuracy of detecting the key states of subgoals. Initially, the agent does not know any of these (hidden) subgoals, and no symbolic observation is available. The agent only knows the result of task accomplishment at the end of the trajectory.}
    \label{fig:contrastive_acc}
\end{figure}

\begin{figure}
    \centering
    \subfigure[Task 1 Letter]{
        \centering
        \includegraphics[width=1.5in]{plot_letter_task1.png}
    }
    \subfigure[Task 2 Letter]{
        \centering
        \includegraphics[width=1.5in]{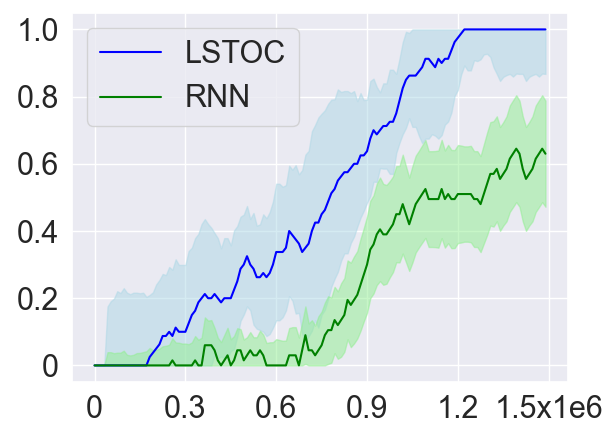}
    }
    \subfigure[Task 3 Letter]{
        \centering
        \includegraphics[width=1.5in]{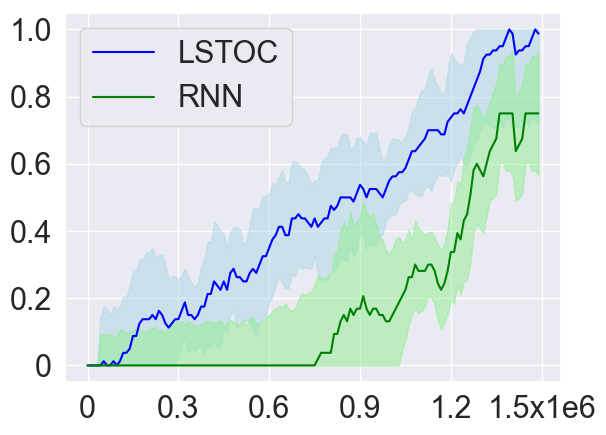}
    }

    \subfigure[Task 1 Office]{
        \centering
        \includegraphics[width=1.5in]{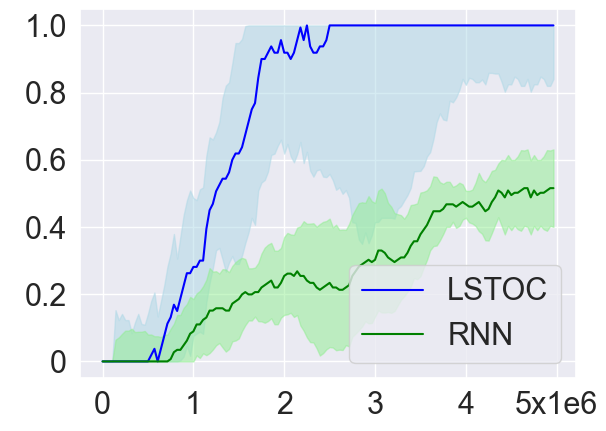}
    }
    \subfigure[Task 2 Office]{
        \centering
        \includegraphics[width=1.5in]{plot_office_task2.png}
    }
    \subfigure[Task 3 Office]{
        \centering
        \includegraphics[width=1.5in]{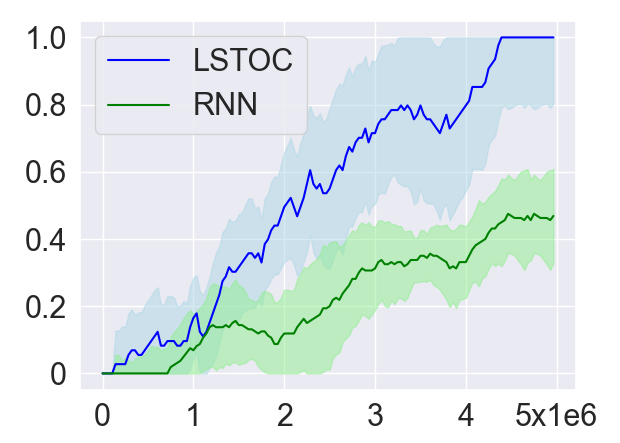}
    }

    \subfigure[Task 1 Crafter]{
        \centering
        \includegraphics[width=1.5in]{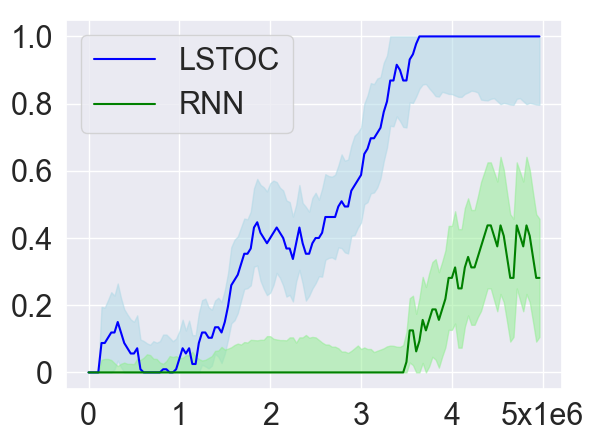}
    }
    \subfigure[Task 2 Crafter]{
        \centering
        \includegraphics[width=1.5in]{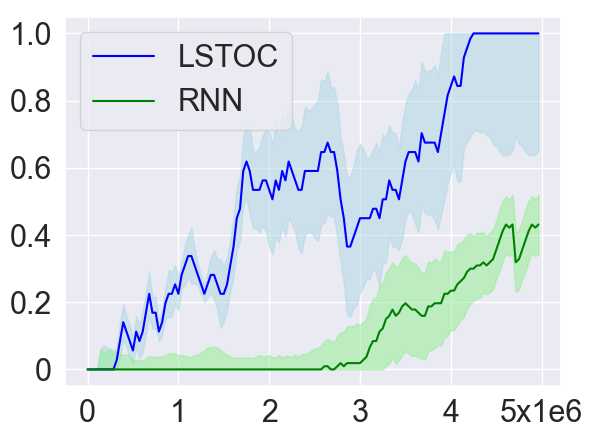}
    }
    \subfigure[Task 3 Crafter]{
        \centering
        \includegraphics[width=1.5in]{plot_minihack_task3.png}
    }
    \caption{Comparison of learning efficiency. The x-axis is the environmental steps taken by the agent. The y-axis is the success rate on completing the given task.} 
    \label{fig:learning_efficiency}
\end{figure}

\section{Neural Architecture}
\label{sec:arch}
We build neural network architectures for state representation function $o_{\theta}$ and importance function $f_{\omega}$ and the exploration policy $\pi_{\text{exp}}$. The function $o_{\theta}$ is used to extract a $d$-dimensional vector as state representation for an input raw state or observation. In Letter and Office domain, $d$ is $128$ and $o_{\theta}$ is realized by a two-layer MLP with 128 neurons in each layer. In Crafter domain, $o_{\theta}$ is realized by a convolutional neural network (CNN) module. This CNN is the same as the classical CNN for deep RL proposed in \citep{mnih2015human}, where the first convolutional layer has 32 channels with kernel size of 8 and stride of 4, the second  layer has 64 channels with the kernel size of 4 and stride of 2 and the third layer has 64 channels with the kernel size of 3 and stride of 1. The CNN module produces an embedding vector with the size of $d=512$.

The importance function $f_{\omega}$ is realized by MLP in all three domains, which is a two-layer MLP with $128$ neurons in Letter and Office domains and $256$ neurons in Crafter domain. 

The exploration policy $\pi_{\text{exp}}$ is a GRU-based policy. In $\pi_{\text{exp}}$, the hidden dimension of gated recurrent unit (GRU) module is 128 for Letter/Office domains and 256 for Crafter domain. The outputs of $\pi_{\text{exp}}$ consist of action and predicted value, which are conditioned on both the hidden state and the embedding vector of input observation.

\section{Correctness and Limitation}
\label{sec:app_correctness_and_limitation}
\subsection{Correctness}
Empirically, as long as $N_T$ is large enough and randomness of exploration policy $\pi_{\text{exp}}$ is sufficient, the contrastive learning in \eqref{con-obj} can always discover the correct key state of next subgoal at every working node. This is because sufficient trajectory data and randomness in data collection can make most parts of state space covered by both positive and negative trajectories. Then, the correct subgoal tree can be built and labeling function can be correctly obtained by solving the ILP problem. Specifically, the randomness of $\pi_{\text{exp}}$ can be guaranteed by using $\epsilon$-greedy into action selection, where $\epsilon=0.5$ is enough for all the environments. 

In other words, whenever the state space is covered sufficiently enough by collected trajectory data, no hidden subgoal in the environment can be missed. Then, whenever every collected positive trajectory can be explained by some path in $\mathcal{T}_{\varphi}$, all the hidden subgoals and their temporal orderings are learned, showing the correctness of the termination condition of LSTOC.

\subsection{Limitation}
\begin{figure}[H]
    \vspace{-5pt}
    \centering
    \includegraphics[width=2.5in]{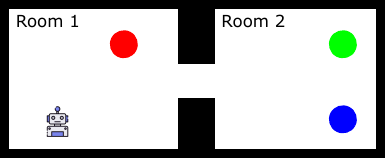}
    \caption{Map of room 1 and 2.}
    \label{fig:limit0}
    \vspace{-10pt}
\end{figure}

However, LSTOC framework still has limitations. First, its labeling component cannot tell the difference between the bottleneck state in the environment and the real key states of subgoals, even though these are all key states. For example, as shown in Figure \ref{fig:limit0}, a navigation environment has two rooms. 

Room 1 has a red ball and room 2 has a blue ball and a green ball. However, there is an open corridor connecting room 1 and 2. This corridor becomes a bottleneck state of connecting room 1 and 2. Assume that the given task has hidden subgoals of first picking up red ball, then blue ball, and finally green ball, i.e., $r;b;g$ in temporal logic language. In this case, four subgoals will be discovered by LSTOC, which are corridor and states of red, blue and green balls. Then, since the FSM does not have corridor and the agent does not know the location of blue ball, the the corridor cannot be distinguished from the state of blue ball. Therefore, the ILP problem in the labeling component does not have a unique solution and the labeling function cannot be obtained. However, the learning subgoal part of LSTOC can still detect every meaningful key state. 

Second, the labeling component cannot tell the differences of symmetric branches of FSM. For example, the given FSM is $(a;b;c)\vee(d;e;f)$ and two branches are symmetric. The learning subgoal component of LSTOC will discover 6 key states of subgoals, but the mapping from discovered key states to semantic symbols (a,b,c,d,e, and f) cannot be determined.

{\bf
It is important to note that the limitations outlined above cannot be resolved. In the problem formulation, the only feedback available to the agent for anchoring subgoal symbols is a binary label indicating task completion at the end of each trajectory. The agent learns the labeling function (i.e., the mapping from learned subgoals to subgoal symbols in the given FSM), by utilizing these binary labels and structural information of subgoal symbols in the given FSM. As a result, when there are hidden bottlenecks in the environment or the task involves symmetric branches of subgoals, the given FSM has ambiguity and hence the agent lacks sufficient information to accurately identify the mapping from learned subgoals to subgoal symbols in the given FSM. Consequently, in these situations, it is impossible for the agent to correctly learn the labeling function, but the agent can still accurately estimate the hidden subgoals in the environment.
}

Third, the trajectory collection could not cover some important states in hard-exploration environments. In environments like Montezuma-revenge, some key states need thousands of actions to reach and the exploration policy trained by task completion signal only cannot collect trajectories covering these key states. Then, some hidden subgoals cannot be discovered and the labeling component cannot produce correct result.

\end{document}